\documentclass[lettersize,journal]{IEEEtran}
\usepackage{amsmath,amsfonts}
\usepackage{amssymb}
\usepackage{algorithm}
\usepackage[noend]{algpseudocode}
\usepackage{array}
\usepackage[caption=false,font=normalsize,labelfont=sf,textfont=sf]{subfig}
\usepackage{textcomp}
\usepackage{stfloats}
\usepackage{url}
\usepackage{verbatim}
\usepackage{graphicx}
\usepackage{cite}
\usepackage{marvosym}
\usepackage{tabularx}
\usepackage{makecell}
\usepackage{colortbl}
\usepackage{booktabs}
\usepackage{multirow}
\usepackage[normalem]{ulem}
\usepackage[caption=false]{subfig}
\usepackage{xcolor}
\usepackage{float}
\usepackage{rotating}
\definecolor{ManBlue}{RGB}{222,240,255}

\newcolumntype{M}{>{\columncolor{ManBlue}}c}
\newcommand{\bb}[1]{\cellcolor{ManBlue}{\textbf{#1}}}
\usepackage[normalem]{ulem}
\usepackage{cancel}

\DeclareRobustCommand{\DEL}[1]{{\color{red}\ifmmode\cancel{#1}\else\sout{#1}\fi}}

\newcommand{\mybio}[3]{%
\par\noindent
\begin{minipage}[t]{1.02in}
\vspace{0pt}%
\includegraphics[width=1in,height=1.25in,clip,keepaspectratio]{#1}
\end{minipage}\hspace{0.15in}%
\begin{minipage}[t]{\dimexpr\columnwidth-1.02in-0.15in\relax}
\vspace{0pt}%
\textbf{#2}~#3
\end{minipage}\par\vspace{3pt}%
}

\begin{document}

\title{MAN++: Scaling Momentum Auxiliary Network for Supervised Local Learning in Vision Tasks}

\author{Junhao Su, Feiyu Zhu, Hengyu Shi, Tianyang Han, Yurui Qiu, Junfeng Luo, Xiaoming Wei, Jialin Gao
\thanks{Junhao Su, Hengyu Shi, Tianyang Han, Yurui Qiu, Junfeng Luo, Xiaoming Wei and Jialin Gao are with Vision Intelligence in MeiTuan, Beijing and Shanghai, China. Email: \{sujunhao02, shihengyu02, hantianyang, qiuyurui, luojunfeng, weixiaoming, gaojialin04\}@meituan.com}
\thanks{Feiyu Zhu is with AttrSense, Shanghai, China. Email: zeurdfish@gmail.com}
\thanks{Junhao Su, Feiyu Zhu, Hengyu Shi did this work with eqaul contributions. Junhao Su was responsible for writing the entire manuscript, designing and conducting the ablation and analysis experiments, and carrying out all experimental evaluations. Feiyu Zhu implemented the method in code and adapted it for application on the COCO and CityScapes datasets. Hengyu Shi created the structural diagrams illustrating the method and its details.}
\thanks{The corrsponding author is Jialin Gao.}
\thanks{A preliminary version of this research has appeared in ECCV 2024 Oral \cite{man}.}
}

\markboth{IEEE Transactions on Pattern Analysis and Machine Intelligence}
{Shell \MakeLowercase{\textit{et al.}}: A Sample Article Using IEEEtran.cls for IEEE Journals}


\maketitle
\begin{abstract}
End-to-end backpropagation remains the dominant training paradigm in deep learning, yet it suffers from inherent drawbacks, including update locking, high GPU memory consumption, and limited biological plausibility.
Supervised local learning alleviates these issues by dividing the network into multiple blocks and training each block independently with an auxiliary network. 
However, gradient isolation also weakens the influence of downstream representations on earlier blocks, often resulting in a clear accuracy gap to end-to-end training.
We propose \textbf{Momentum Auxiliary Network++ (MAN++)}, a scalable framework that improves supervised local learning via a lightweight parameter-space transfer between adjacent blocks. MAN++ employs the \textbf{exponential moving average (EMA)} of parameters from adjacent blocks to propagate contextual information across the network. To address feature mismatches arising from direct EMA parameter transfer, we introduce a \textbf{learnable scaling bias}, which compensates feature statistics mismatch and stabilizes the transfer.
Extensive experiments on image classification, object detection, and semantic segmentation across multiple architectures illustrate that MAN++ achieves accuracy on par with end-to-end training while substantially reducing GPU memory usage. These results position MAN++ as a practical and effective alternative to conventional backpropagation, offering new insights into scalable supervised local learning for vision tasks.
\end{abstract}

\section{Introduction}
\label{sec:intro}

In traditional deep learning, enhancing performance is typically achieved by increasing network depth, which in turn renders end-to-end backpropagation indispensable for model training \cite{2}.
However, as the depth of the network increases, so does the computational cost associated with evaluating the loss function and performing continuous gradient descent through successive layers to optimize parameters \cite{20, 21, 14}. 
Moreover, parameter updates occur only after the complete forward and backward propagation passes, imposing a "locking" constraint on the updates \cite{1}. 
This updating scheme, coupled with the locking issue, stands in stark contrast to the localized signal processing observed in biological synapses \cite{7,35,36} and exacerbates challenges such as reduced parallelism and increased GPU memory consumption \cite{16,18}, ultimately affecting training efficiency and scalability.
\begin{figure*}[!htbp]
    \centering
    \includegraphics[width=\linewidth]{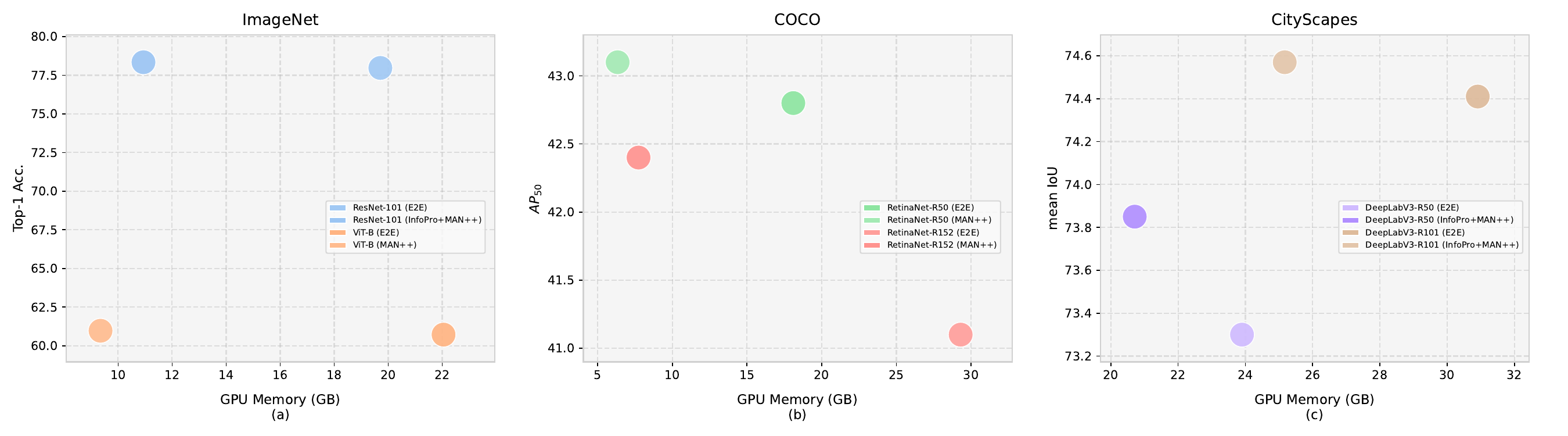}
    \caption{Comparison of accuracy across different datasets and backbones for both MAN++ and E2E methods. Fig~\ref{Figure 1}.(a) shows the results of training from scratch on the ImageNet dataset for 90 epochs. Fig~\ref{Figure 1}.(b) presents the results of training on the COCO dataset for 100 epochs using pretrained weights from ImageNet. Fig~\ref{Figure 1}.(c) displays the results of training on the CityScapes dataset for 4,000 iterations, also using pretrained weights from ImageNet.}
    \label{Figure 1}
\end{figure*}

To address these challenges, an alternative training paradigm, namely local learning, has recently emerged \cite{1,11,12,13,14,15,27,28,43,mlaan}.
Unlike end-to-end training, local learning partitions the neural network into multiple gradient-isolated blocks, which independently perform backpropagation, thereby preventing gradient flow between them.
The parameters of each module are updated by its own auxiliary network, driven by distinct local objectives \cite{15,27,28}. 
This approach mitigates the locking issue by enabling each gradient-isolated module to update its parameters immediately upon receiving local error signals, thereby avoiding the sequential update bottleneck of end-to-end training and significantly enhancing parallel training efficiency \cite{45,46}.
Furthermore, local learning retains only the gradients of each module’s backbone and auxiliary networks during training and promptly releases them after local updates. It substantially reduces GPU memory requirements and obviates the need to store extensive global gradient information \cite{16,18}.

Nevertheless, while local learning alleviates the locking problem and conserves GPU memory, a significant performance gap relative to end-to-end training persists, precluding its complete replacement.
Existing techniques in local learning primarily focus on refining the structure of the auxiliary network \cite{28} and narrowing the performance gap by enhancing the local loss function \cite{15,27}.
However, these improvements do not fully address a core limitation of gradient isolation: earlier blocks are optimized mainly by local signals and receive only indirect feedback about how their representations will be used by later blocks.
This mismatch becomes more pronounced when the network is split into many blocks, where locally optimal features may not compose well into a strong final predictor.

In this paper, we propose a novel local learning network architecture—Momentum Auxiliary Network++ (MAN++).
MAN++ introduces a lightweight momentum-based parameter transfer to better couple adjacent blocks during supervised local learning.
Specifically, MAN++ comprises two components: an EMA module and a Scale Learnable Bias module, both positioned between the main network and the auxiliary network of each gradient-isolated module. 
The EMA module employs the Exponential Moving Average (EMA) technique \cite{34} to integrate parameters from the subsequent module into its update process. This innovative approach enables each local module to incorporate information beyond its immediate local objective, thereby promoting a closer alignment with the network’s global goal.

However, directly applying EMA parameters for updates introduces challenges, such as feature statistics mismatch between gradient-isolated blocks and the slow adaptation of purely momentum-based transfer.
To overcome these issues, we introduce a Scale Learnable Bias that augments the EMA module’s updates, enhancing the effective sharing of information.
Notably, the proposed MAN++ incurs only a minimal increase in GPU memory usage while delivering significant performance improvements.
We validate MAN++ through experiments on various CNN and ViT architectures across multiple datasets for image classification, object detection, and semantic segmentation.
The experimental results demonstrate that MAN++ significantly narrows the gap to end-to-end training under local learning, achieving performance comparable to end-to-end training and offering a promising alternative.

The contributions of this paper can be summarized as follows:
\begin{itemize}
\item We propose Momentum Auxiliary Network++ (MAN++), which introduces an EMA-based momentum transfer with a learnable scale-and-bias compensation for supervised local learning.
\item MAN++ is a plug-and-play method that can be seamlessly integrated into any supervised local learning framework, significantly broadening its applicability while imposing only minimal additional GPU memory overhead.
\item MAN++ demonstrates its effectiveness across different visual tasks using various network architectures, achieving state-of-the-art performance. It significantly reduces memory usage while matching the performance of end-to-end training methods, thus providing a potential alternative to end-to-end training for local learning.
\end{itemize}
\begin{figure*}[t]
    \centering
    \includegraphics[width=\textwidth]{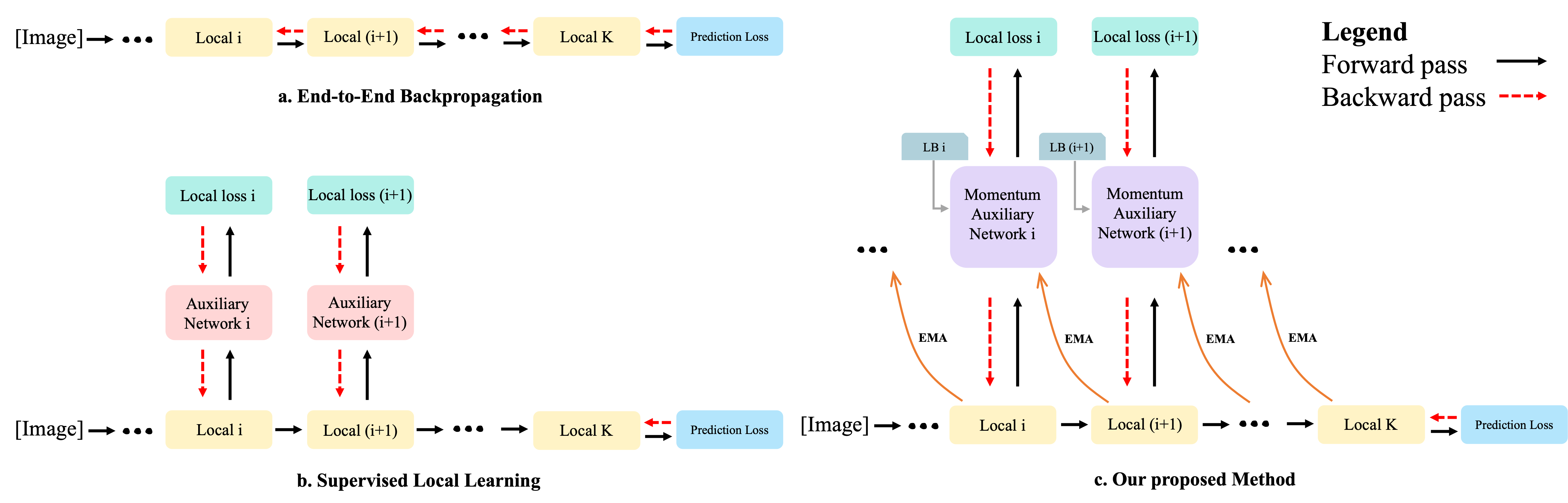}
    \caption{Comparison of (a) end-to-end backpropagation, (b) other supervised local learning methods, and (c) our proposed method. Unlike E2E, supervised local learning separates the network into K gradient-isolated local blocks. LB stands for the Learnable Bias.}
    \label{Figure 2}
\end{figure*}
\section{Related Work}
\label{sec:formatting}
\subsection{Local Learning} Local learning was initially proposed as an alternative to end-to-end training, offering significant advantages in terms of GPU memory savings. From a biological plausibility perspective, block-based learning models are also closer to the way the human brain learns \cite{9}. This approach notably reduces the heavy reliance of deep neural networks on computational resources, alleviating the issue of update locking that arises in end-to-end training \cite{41}, and has driven the development of alternative supervised local learning methods. Examples of these methods include a differentiable search algorithm that divides the network into multiple gradient-isolated blocks for learning, with a manually designed auxiliary network chosen for each local block \cite{42}, or a self-supervised contrastive loss function that adheres to the training principles of local learning \cite{13,43}. In the current field of local learning \cite{glcan,man,hpff,lakd}, improvements mainly focus on manually designing better auxiliary networks and developing more effective local losses.

\noindent {\bfseries Supervised Local Loss:} PredSim \cite{15} designed a layer-wise supervised loss for local learning, consisting of two loss functions: cross-entropy loss and similarity matching loss. The similarity matching loss is used to evaluate the similarity between layers to ensure that important features are not lost. These two losses are computed only between gradient-isolated local hidden layers, eliminating the need to propagate global errors back to the hidden layers. These methods have achieved state-of-the-art performance in the field of supervised local learning. InfoPro \cite{27} further introduced a local reconstruction loss, which measures the restoration of features from intermediate hidden layers by comparing them with the input image, preserving some of the original information in the intermediate layers. This helps prevent the loss of globally beneficial features due to an excessive focus on local objectives. While these approaches have brought significant progress to local learning, there remains a noticeable gap compared to the performance of end-to-end training.

\noindent {\bfseries Auxiliary Networks:} DGL \cite{28} designed a simple auxiliary network consisting of three consecutive convolutional layers connected to a global pooling layer, followed by three consecutive fully connected layers. This simplified structure limits the auxiliary network's parameters to only 5\% of the main network's capacity, offering advantages in terms of speed and further savings in GPU memory. InfoPro \cite{27} designed two different auxiliary networks. One network consists of a convolutional layer followed by two fully connected layers, while the other network uses two convolutional layers along with an upsampling operation. Auglocal \cite{auglocal} designs an auxiliary network for each local network block based on the structure of the network, ensuring that the structure of the auxiliary network aligns with some subsequent layers. This approach eliminates the need for manually designing auxiliary networks and ensures better coupling between the local network blocks and the overall network. MLAAN \cite{mlaan} introduces part of the structure and parameters of subsequent layers as auxiliary networks using the EMA method, and combines local network blocks locally by partitioning, allowing some local network blocks to share weights. DGL and InfoPro tend to exhibit poor performance when the network is divided into a large number of local network blocks. In contrast, while Auglocal and MLAAN demonstrate superior performance, their heavyweight auxiliary network designs result in slower training speeds and, compared to end-to-end training, the advantages in GPU memory usage are less significant.

\subsection{Alternative Learning Rules to E2E Training}
The inherent limitations of end-to-end training have sparked growing attention in recent years towards exploring alternative methods to E2E training \cite{10}. Additionally, some recent studies have attempted to completely avoid using backpropagation in neural networks through forward gradient learning \cite{54,55}. At the same time, the issue of weight transmission \cite{9} has been addressed by employing different feedback connections \cite{47,48} or directly propagating global errors to each hidden unit \cite{49,50}. Replacement learning \cite{replace} achieves this by synthesizing the parameters of certain layers with those of the preceding and succeeding layers, allowing these layers to bypass backpropagation. Although these methods alleviate some inherent flaws of E2E training, such as biological implausibility, to some extent, they still rely on global objectives, which fundamentally differs from biological neural networks, where information is transmitted through local synapses in the human brain. Additionally, due to various limitations, these methods do not perform satisfactorily on large datasets.
\section{Method}

\begin{table*}[t]
  \centering
  \scriptsize
  \caption{Notation used in Sec.~Method and Sec.~Experiments.}
  \label{tab:notation_main}
  \setlength{\tabcolsep}{6pt}
  \renewcommand{\arraystretch}{1.15}
  \begin{tabular}{l p{0.72\textwidth}}
    \toprule
    \textbf{Symbol} & \textbf{Meaning} \\
    \midrule
    $x,\,y$ &
    Input sample and its ground-truth label. \\
    $\hat{y}$ &
    Final prediction produced by the last block / output layer. \\
    $\mathcal{L}(\hat{y},y)$ &
    Supervised loss between prediction and label. \\

    \midrule
    $K$ &
    Number of gradient-isolated local blocks after partitioning the backbone. \\
    $j \in \{1,\dots,K\}$ &
    Index of a local block; auxiliary heads are attached to blocks $j\in\{1,\dots,K-1\}$. \\
    $x_j$ &
    Input activation to block $j$ in the forward pass; $x_1=x$. \\
    $x_{j+1}=f_{\theta_j}(x_j)$ &
    Forward mapping of block $j$; $f_{\theta_j}(\cdot)$ denotes the computation of block $j$. \\
    $\theta_j$ &
    Parameters of the $j$-th local block. \\

    \midrule
    $g_{\gamma_j}(\cdot)$ &
    Auxiliary network / head attached to block $j$ (for $j<K$). \\
    $\hat{y}_j=g_{\gamma_j}(x_{j+1})$ &
    Local prediction produced by the $j$-th auxiliary head. \\
    $\gamma_j$ &
    Parameters of the $j$-th auxiliary network. \\
    $b_j$ &
    Learnable bias term in the $j$-th auxiliary network (used in MAN++). \\
    $s_j$ &
    Learnable scalar that controls the strength of EMA-based transfer in block $j$
    (by default, block-specific and shared within the block). \\

    \midrule
    $\eta_l,\,\eta_a$ &
    Learning rates for local blocks and auxiliary networks, respectively. \\
    $\alpha$ &
    EMA decay (default $\alpha=0.995$ unless otherwise stated). \\
    $EMA(\cdot)$ &
    Exponential Moving Average operator used for parameter transfer from block $(j+1)$ to block $j$. \\
    $\mathrm{detach}(\cdot)$ &
    Stop-gradient operation used to break backpropagation across blocks (e.g., in PPLL). \\

    \midrule
    \texttt{DDP}, \texttt{PPLL} &
    DistributedDataParallel and Pipeline Parallel Local Learning, respectively. \\
    \texttt{PPLL with GPUs}$=K$ &
    The setting where each block (and its head, if any) is placed on one GPU. \\
    $\mathcal{D}$, $E$ &
    Dataset and the number of training epochs (Algorithm~\ref{alg:ppll_manpp}). \\
    $L_i,\,H_i$ &
    In PPLL pseudocode, the $i$-th local block and its auxiliary head (if $i<K-1$). \\
    $\texttt{inQ}[i],\,\texttt{outQ}[i]$ &
    Message queues that pass activations between GPU stages in PPLL. \\
    \bottomrule
  \end{tabular}
\end{table*}

\subsection{Preliminaries}
To provide context, we first review traditional end-to-end supervised learning and the backpropagation mechanism.
Let $x$ denote an input sample with the corresponding ground truth label $y$.
The entire deep network is partitioned into multiple local blocks. In the forward pass, the output of the $j$-th block is used as the input to the 
($j+1$)-th block, \textit{i.e.}, $x_{j+1}=f_{\theta_j}(x_j)$, where $\theta_{j}$ represents the parameters of the $j$-th local block and $f(\cdot)$ denotes its forward computation.
The loss function ${\mathcal{L}}(\hat{y}, y)$ is calculated between the output of the final block $\hat{y}$ and the ground truth $y$, and this loss is then backpropagated to update the preceding blocks.

Supervised local learning \cite{15,27,28} introduces local supervision by incorporating auxiliary networks. For each gradient-isolated local block, an auxiliary network is attached. 
The output from the local block is fed into its corresponding auxiliary network, which generates a local supervision signal: $\hat{y_{j}}=g_{\gamma_j}(x_{j+1})$. Here, $\gamma_{j}$ denotes the parameters of the $j$-th auxiliary network.

In this setup, the parameters of the $j$-th auxiliary network and the local block, $\gamma_j, \theta_j$ respectively, are updated as follows:

\begin{equation}
\gamma_j \leftarrow \gamma_j - \eta_a \times \nabla_{\gamma_j} \mathcal{L}(\hat{y_j}, y)
\end{equation}

\begin{equation}
\theta_j \leftarrow \theta_j - \eta_l \times \nabla_{\theta_j} \mathcal{L}(\hat{y_j}, y)
\end{equation}

\noindent where $\eta_a, \eta_l$ represent the learning rates for the auxiliary network and the local block, respectively. By attaching an auxiliary network, each local block becomes gradient-isolated and can be updated through local supervision, thereby circumventing the need for global backpropagation.

\subsection{Momentum Auxiliary Network++}
Current approaches typically integrate supervision signals into each local block, enabling parallel parameter updates and lower memory overhead. However, such methods may suffer from a myopic problem, where each local block fails to incorporate information from subsequent blocks, ultimately leading to suboptimal final accuracy.

To mitigate this issue, we introduce a comprehensive dynamic information exchange module, Momentum Auxiliary Network++ (MAN++). MAN++ leverages an Exponential Moving Average (EMA) mechanism \cite{34} as an information transfer channel to propagate data from subsequent blocks back to the current block. In the MAN++ framework, the parameters of the $j$-th auxiliary network and the corresponding local block are updated as follows:

\begin{equation}
\gamma_j \leftarrow \gamma_j - \eta_a \times \nabla_{\gamma_j} \mathcal{L}(\hat{y_j}, y)
\end{equation}

\begin{equation}
\gamma_j \leftarrow s_{j} \times EMA(\gamma_j, \theta_{j+1})
\end{equation}

\begin{equation}
\theta_j \leftarrow \theta_j - \eta_l \times \nabla_{\theta_j} \mathcal{L}(\hat{y_j}, y)
\end{equation}

\noindent where $\gamma_j$ denotes the parameters of the $j$-th auxiliary networ, $\theta_j$ represents those of the local block $j$, and $s_{j}$ is a learnable parameter used to control the strength of the EMA-based transfer in block $j$.
Following updates of the local gradient, $\gamma_j$ is further optimized via an EMA update using parameters from the subsequent block, i.e., a weighted moving average in parameter space.

\begin{figure}[t]
    \centering
    \includegraphics[width=\linewidth]{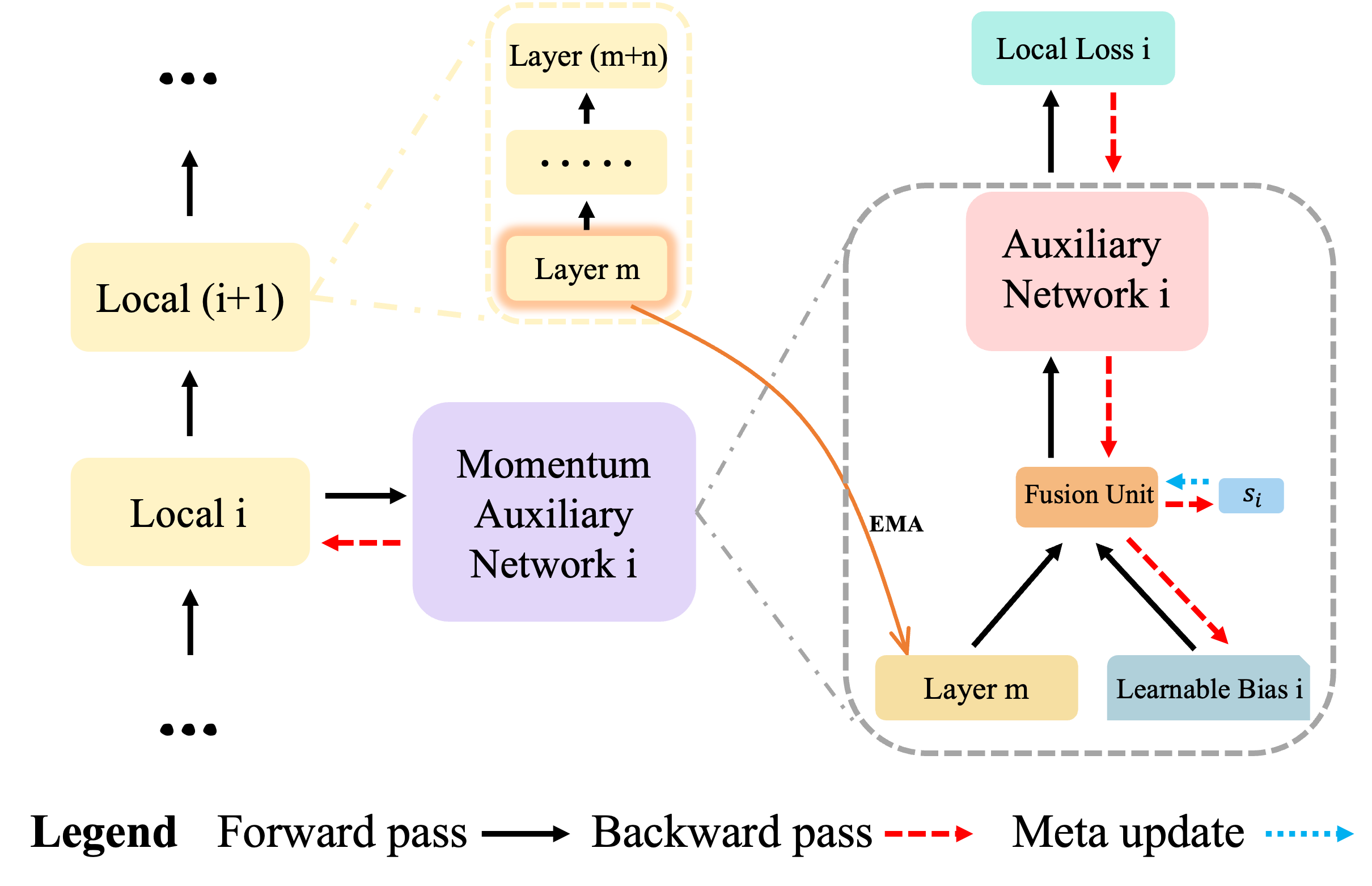}
    \caption{Details of the Momentum Auxiliary Network. Local (i+1) represents the (i+1)-th gradient-isolated local block, which contains layers from layer m to layer (m+n), totaling n+1 layers (n$\geqslant$0). We only use the parameters of the first layer to ensure a balance in GPU memory usage. Specifically, for ResNet, “first layer” refers to the first residual unit (i.e., a basic block or bottleneck), whereas for Vision Transformer, “first layer” refers to the first Transformer block.}
    \label{Figure 3}
\end{figure} 

In our experiments, we observed that directly updating the auxiliary network with EMA parameters yielded only marginal improvements for each local block.
Analysis revealed that differences in the features learned by each local block hinder the EMA update, which, being gradient-free, is not capable of adaptive optimization.
To overcome this limitation, we introduced a learnable bias and employed the learnable parameter $s_j$ to balance the contributions from the EMA and the bias. This modification enhances the learning capacity of the hidden layers within the local blocks and compensates for the deficiencies of the standalone EMA update.

The parameter update for the $j$ auxiliary network is thus formulated as follows:

\begin{equation}
\label{eq.6}
(\gamma_j,b_j) \leftarrow (\gamma_j,b_j) - (2-s_j) \times \eta_a \times \nabla_{(\gamma_j,b_j)} \mathcal{L}(\hat{y_j}, y)
\end{equation}

\begin{equation}
\label{eq.7}
\gamma_j \leftarrow s_j \times EMA(\gamma_j, \theta_{j+1})
\end{equation}

\noindent where ($\gamma_j,b_j$) collectively represents the parameters of the $j$-th auxiliary network. In this process, $\gamma_j$ is updated jointly via the learnable bias $b_j$ and the EMA, with their respective contributions balanced by $s_j$.

\textbf{Architecture-specific instantiation: } In Eq.~\ref{eq.7}, the EMA source term uses parameters from the first computational module of the next local block $(j{+}1)$ to control memory overhead.
For ResNet-style CNNs, this corresponds to the first residual unit (\texttt{BasicBlock} or \texttt{Bottleneck}) in local block $(j{+}1)$.
For ViT backbones, each local block consists of several Transformer blocks; we use the first Transformer block in local block $(j{+}1)$, and instantiate the EMA source parameters using its attention and feed-forward components, \emph{i.e.}, the QKV projection in the MHA module together with the MLP/FFN submodule.

\textbf{Granularity of the learnable scale/bias: } By default, both the learnable scale $s_j$ and the learnable bias $b_j$ are local block-specific: each gradient-isolated local block $j$ has its own independent $(s_j,b_j)$ with no sharing across local blocks.
We parameterize $s_j$ as a single scalar per local block (applied at the local block level).
The bias $b_j$ is part of the $j$-th auxiliary network parameters and is optimized jointly with $\gamma_j$ via Eq.~\ref{eq.6}, while the EMA-based information transfer is applied via Eq.~\ref{eq.7}.

Overall, MAN++ augments standard supervised local learning with (i) an EMA-based parameter transfer from the subsequent block and (ii) a learnable scale-and-bias term that compensates for feature statistics mismatch during transfer.
We demonstrate in Sec.~X that this design is compatible with diverse backbones and vision tasks with minimal overhead.

\section{Experiments}

\subsection{Experimental Setup}
We conducted experiments on six datasets covering three different types of tasks: classification, detection, and segmentation. The datasets include \textbf{CIFAR-10}~\cite{25}, \textbf{SVHN}~\cite{30}, \textbf{STL-10}~\cite{31}, \textbf{ImageNet}~\cite{32}, \textbf{COCO}~\cite{coco}, and \textbf{Cityscapes}~\cite{cityscapes}, using ResNets~\cite{24} and Vision Transformers~\cite{vit} of varying depths as the network architectures.

We selected two state-of-the-art supervised local learning methods for comparison, namely \textbf{DGL}~\cite{28}, \textbf{InfoPro}~\cite{27}. Specifically, the networks are divided into $K$ local blocks, each containing approximately the same number of layers. Our proposed \textbf{Momentum Auxiliary Network++} is applied only to the first $K{-}1$ local blocks, while the $K$-th block does not use any auxiliary network and connects directly to the output layer.

We compare this configuration with both the conventional end-to-end (E2E) training method and the original supervised local learning baselines, ensuring consistent training settings to eliminate confounding factors that may affect the experimental results.

\begin{table*}[!htbp]
  \centering
  \caption{Performance of different networks with varying numbers of local blocks. “+MAN” denotes adding MAN, “+MAN++” denotes the enhanced MAN++. The average test error is obtained from 5 runs.}
  \resizebox{\textwidth}{!}{
  \begin{tabular}{c|ccccc}\hline
    \multirow{2}{*}{Dataset} & \multirow{2}{*}{Method} & \multicolumn{2}{c}{ResNet-32} & \multicolumn{2}{c}{ResNet-110} \\\cline{3-6}
                             &                         & K = 8 (Test Error) & K = 16 (Test Error) & K = 32 (Test Error) & K = 55 (Test Error) \\ \hline
    \multirow{6}{*}{\makecell[c]{CIFAR-10 \\ (E2E(ResNet-32)=6.37, \\ E2E(ResNet-110)=5.42)}} 
       & DGL                  & 11.63 & 14.08 & 12.51 & 14.45 \\
       & DGL+MAN              &  8.42($\downarrow$3.21) &  9.11($\downarrow$4.97) &  9.65($\downarrow$2.86) &  9.73($\downarrow$4.72) \\
       & \bb{\textbf{DGL+MAN++}}   & \bb{\textbf{8.06($\downarrow$3.57)}} & \bb{\textbf{8.75($\downarrow$5.33)}} & \bb{\textbf{9.05($\downarrow$3.46)}} & \bb{\textbf{9.51($\downarrow$4.94)}} \\
       & InfoPro              & 11.51 & 12.93 & 12.26 & 13.22 \\
       & InfoPro+MAN          &  9.32($\downarrow$2.19) &  9.65($\downarrow$3.28) &  9.06($\downarrow$3.20) &  9.77($\downarrow$3.45) \\
       & \bb{\textbf{InfoPro+MAN++}} & \bb{\textbf{7.97($\downarrow$3.54)}} & \bb{\textbf{8.25($\downarrow$4.68)}} & \bb{\textbf{7.75($\downarrow$4.51)}} & \bb{\textbf{8.67($\downarrow$4.55)}} \\ \hline
    \multirow{6}{*}{\makecell[c]{STL-10 \\ (E2E(ResNet-32)=19.35, \\ E2E(ResNet-110)=19.67)}} 
       & DGL                  & 25.05 & 27.14 & 25.67 & 28.16 \\
       & DGL+MAN              & 20.74($\downarrow$4.31) & 21.37($\downarrow$5.77) & 22.54($\downarrow$3.13) & 22.69($\downarrow$5.47) \\
       & \bb{\textbf{DGL+MAN++}}   & \bb{\textbf{20.17($\downarrow$4.88)}} & \bb{\textbf{21.08($\downarrow$6.06)}} & \bb{\textbf{20.73($\downarrow$4.94)}} & \bb{\textbf{21.57($\downarrow$6.59)}} \\
       & InfoPro              & 27.32 & 29.28 & 28.58 & 29.20 \\
       & InfoPro+MAN          & 23.17($\downarrow$4.15) & 23.54($\downarrow$5.74) & 24.08($\downarrow$4.50) & 24.74($\downarrow$4.46) \\
       & \bb{\textbf{InfoPro+MAN++}} & \bb{\textbf{21.14($\downarrow$6.18)}} & \bb{\textbf{20.81($\downarrow$8.47)}} & \bb{\textbf{21.73($\downarrow$6.85)}} & \bb{\textbf{23.77($\downarrow$5.43)}} \\ \hline
    \multirow{6}{*}{\makecell[c]{SVHN \\ (E2E(ResNet-32)=2.99, \\ E2E(ResNet-110)=2.92)}} 
       & DGL                  &  4.83 &  5.05 &  5.12 &  5.36 \\
       & DGL+MAN              &  3.80($\downarrow$1.03) &  4.04($\downarrow$1.01) &  4.08($\downarrow$1.04) &  4.52($\downarrow$0.84) \\
       & \bb{\textbf{DGL+MAN++}}   & \bb{\textbf{3.34($\downarrow$1.49)}} & \bb{\textbf{3.83($\downarrow$1.22)}} & \bb{\textbf{3.25($\downarrow$1.87)}} & \bb{\textbf{2.72($\downarrow$2.64)}} \\
       & InfoPro              &  5.61 &  5.97 &  5.89 &  6.11 \\
       & InfoPro+MAN          &  4.49($\downarrow$1.12) &  5.19($\downarrow$0.78) &  4.85($\downarrow$1.04) &  4.99($\downarrow$1.12) \\
       & \bb{\textbf{InfoPro+MAN++}} & \bb{\textbf{3.91($\downarrow$1.70)}} & \bb{\textbf{4.37($\downarrow$1.60)}} & \bb{\textbf{4.51($\downarrow$1.38)}} & \bb{\textbf{4.27($\downarrow$1.84)}} \\ \hline
  \end{tabular}}
  \label{Table.1}
\end{table*}

\subsection{Experimental Setting Details.}
\noindent\textbf{EMA decay.}
Unless otherwise stated, we use a fixed EMA decay $\alpha=0.995$ in all experiments.

\noindent \textbf{CIFAR-10}~\cite{25}, \textbf{SVHN}~\cite{30}, and \textbf{STL-10}~\cite{31}.  
    All experiments were conducted on a single Nvidia A100 using \textbf{ResNet-32} and \textbf{ResNet-110}~\cite{24} backbones.  
    We adopted SGD with Nesterov momentum (momentum~=~0.9, weight decay~=~$1\times10^{-4}$).  
    Batch sizes were 1024 for CIFAR-10 and SVHN, and 128 for STL-10.  
    Models were trained for 400~epochs. Initial learning rates were 0.8 for CIFAR-10 and SVHN, and~0.1 for STL-10.  
    A cosine‐annealing schedule~\cite{31} was employed.

\noindent \textbf{ImageNet}~\cite{32}.  
    Training was performed on 8 Nvidia A100 GPUs with the following configurations:  
    \begin{itemize}
        \item \textbf{VGG13}~\cite{33}: 90~epochs, initial LR~=~0.2, batch size~=~512.  
        \item \textbf{ResNet-101} and \textbf{ResNet-152}~\cite{24}: 90~epochs, initial LR~=~0.4, batch size~=~1024.  
        \item \textbf{ResNeXt-101, 32$\times$8d}~\cite{xie2017aggregated}: 90~epochs, initial LR~=~0.2, batch size~=~512.
    \end{itemize}
    Other hyperparameters followed those of the CIFAR-10 setup.

\noindent \textbf{ViT Experiments}.  
    We used \textbf{ViT-Tiny}, \textbf{ViT-Small}, and \textbf{ViT-Base}~\cite{vit}.  
    Training employed the Adam optimizer with a batch size of 1024, an initial learning rate of $1\times10^{-3}$, and lasted for 100~epochs.

\noindent \textbf{COCO}~\cite{coco}.  
    Using 8 Nvidia A100 GPUs and the Adam optimizer, the batch size was 64 and the learning rate was $4\times10^{-4}$.  
    Two settings were evaluated:  
    \begin{enumerate}
        \item Training from scratch (no ImageNet-1K pre-training).  
        \item Training with ImageNet-1K pre-trained weights\footnote{For MAN++, we used weights obtained from MAN++ models trained on ImageNet-1K, reflecting the differences between Local Learning and E2E models.}.
    \end{enumerate}
    Both runs were trained for 100~epochs.

\noindent \textbf{Cityscapes}~\cite{cityscapes}.  
    Experiments used 8 Nvidia A100 GPUs with SGD (momentum~=~0.9, weight decay~=~$1\times10^{-4}$).  
    The batch size was 128, the learning rate was 0.8, and the crop size was 768.  
    As with COCO, two configurations were tested (scratch vs.\ ImageNet-1K pre-training, with MAN++ weights treated identically).  
    Training was carried out for 4k~iterations.

\subsection{Results on Image Classification Datasets}

\noindent \textbf{Results on Image Classification Benchmarks:} We reassess the accuracy of local learning training on \textbf{CIFAR-10} \cite{25}, \textbf{SVHN} \cite{30}, and \textbf{STL-10} \cite{31} with the enhanced \textbf{MAN++} module (Table~\ref{Table.1}).  As before, we split ResNet-32 \cite{24} into $K{=}8,16$ local blocks and ResNet-110 \cite{24} into $K{=}32,55$ blocks.

\textbf{CIFAR-10.} Compared with the original \textbf{DGL} and \textbf{InfoPro} methods, incorporating MAN++ reduces the test error by \textbf{approximately 31-38\%} on the shallow network ResNet-32 (\(K{=}8, 16\)); for example, 11.63$\rightarrow$8.06 and 12.93$\rightarrow$8.25. Even in the deeper ResNet-110 setting (\(K{=}32, 55\)), where global information becomes more critical, MAN++ still yields a \textbf{27-35\%} relative performance improvement, further demonstrating its effectiveness.

\textbf{STL-10.} On this more challenging and lower‐resolution dataset, MAN++ still significantly outperforms all baseline methods. Taking \textbf{DGL} as an example, the test error drops from 27.14 to 21.08 in the ResNet-32 (\(K{=}16\)) setting and from 28.16 to 21.57 in the ResNet-110 (\(K{=}55\)) setting, yielding an \textbf{approximately 22-24\%} performance improvement. For \textbf{InfoPro}, we observe a comparable gain of \textbf{19-29\%}, further confirming that MAN++ does not rely on a specific local training algorithm and possesses strong generalizability.

\textbf{SVHN.} In this digit recognition task, MAN++ achieves the largest relative error reduction. For \textbf{DGL}, the error decreases by \textbf{30-49\%}, dropping from 5.36 to 2.72 in the ResNet-110 (\(K{=}55\)) setting; for \textbf{InfoPro}, the error is reduced by \textbf{approximately 23-31\%}. These substantial improvements indicate that MAN++ can effectively compensate for the lack of global feedback information that commonly occurs in very deep local learning configurations.

\textbf{From MAN to MAN++.} Across all datasets and partition depths, MAN++ delivers an additional \textbf{0.2-1.8 percentage points} of absolute improvement over the already strong MAN baseline (e.g., error drops from 9.11 to 8.75 on CIFAR-10 with ResNet-32, \(K{=}16\), and from 22.54 to 20.73 on STL-10 with ResNet-110, \(K{=}32\)). Hence, the scale momentum auxiliary network introduced in MAN++ further narrows the performance gap with conventional end-to-end (E2E) training, effectively mitigating the long–standing performance disadvantage of supervised local learning methods.

\noindent \textbf{Results on ImageNet:} We further validate the effectiveness of our approach on ImageNet \cite{32} using four networks of varying depths (ResNets \cite{24} and VGG13 \cite{33}). As depicted in Table \ref{Table 2}, when we employ VGG13 as the backbone and divide the network into 10 blocks, DGL \cite{28} achieves merely a Top1-Error of 35.60 and a Top5-Error of 14.2, representing a substantial gap when compared to the E2E method. However, with the introduction of our MAN++, the Top1-Error reduces by 4.67 points, and the Top5-Error decreases by 3.96 points for DGL. This significant enhancement brings the performance closer to the E2E method.

As illustrated in Table \ref{Table 2}, when we use ResNet-101 \cite{24}, ResNet-152 \cite{24}, ResNeXt-101, 32$\times$8d \cite{xie2017aggregated} as backbones and divide the network into four blocks, the performance of InfoPro \cite{27} is already below that of E2E. After incorporating our MAN, the Top-1 Error of these three backbone networks can be reduced by approximately 6\% compared to the original, surpassing the performance of E2E training. These results underscore the effectiveness of our MAN on the large-scale ImageNet \cite{32} dataset, even when using deeper networks.
\begin{table}[!htbp]
\tabcolsep 0.5cm
\scriptsize
\centering
\caption{Results on the validation set of ImageNet}
\scalebox{0.82}{
\begin{tabular}{cccc}\hline
Network & Method & Top1-Error & Top5-Error \\ \hline
\multirow{7}{*}{\makecell[c]{ResNet-101}} 
  & E2E & 22.03 & 5.93 \\
  & InfoPro (K=2)~\cite{27} & 21.85 & 5.89 \\
  & InfoPro+MAN (K=2) & 21.65 ($\downarrow$0.20) & 5.49 ($\downarrow$0.40) \\
  & \bb{\textbf{InfoPro+MAN++ (K=2)}} & \bb{\textbf{21.54 ($\downarrow$0.31)}} & \bb{\textbf{5.41 ($\downarrow$0.48)}} \\
  & InfoPro (K=4)~\cite{27} & 22.81 & 6.54 \\
  & InfoPro+MAN (K=4) & 21.73 ($\downarrow$1.08) & 5.81 ($\downarrow$0.73) \\
  & \bb{\textbf{InfoPro+MAN++ (K=4)}} & \bb{\textbf{21.66 ($\downarrow$1.15)}} & \bb{\textbf{5.72 ($\downarrow$0.82)}} \\ \hline
\multirow{7}{*}{\makecell[c]{ResNet-152}} 
  & E2E & 21.60 & 5.92 \\
  & InfoPro (K=2)~\cite{27} & 21.45 & 5.84 \\
  & InfoPro+MAN (K=2) & 21.23 ($\downarrow$0.22) & 5.53 ($\downarrow$0.31) \\
  & \bb{\textbf{InfoPro+MAN++ (K=2)}} & \bb{\textbf{20.81 ($\downarrow$0.64)}} & \bb{\textbf{5.41 ($\downarrow$0.43)}} \\
  & InfoPro (K=4)~\cite{27} & 22.93 & 6.71 \\
  & InfoPro+MAN (K=4) & 21.59 ($\downarrow$1.34) & 5.89 ($\downarrow$0.82) \\
  & \bb{\textbf{InfoPro+MAN++ (K=4)}} & \bb{\textbf{21.28 ($\downarrow$1.65)}} & \bb{\textbf{5.72 ($\downarrow$0.99)}} \\ \hline
\multirow{7}{*}{\makecell[c]{ResNeXt-101,\\32 $\times$ 8d}} 
  & E2E & 20.64 & 5.40 \\
  & InfoPro (K=2)~\cite{27} & 20.35 & 5.28 \\
  & InfoPro+MAN (K=2) & 20.11 ($\downarrow$0.24) & 5.18 ($\downarrow$0.10) \\
  & \bb{\textbf{InfoPro+MAN++ (K=2)}} & \bb{\textbf{19.84 ($\downarrow$0.51)}} & \bb{\textbf{5.01 ($\downarrow$0.27)}} \\
  & InfoPro (K=4)~\cite{27} & 21.69 & 6.11 \\
  & InfoPro+MAN (K=4) & 20.37 ($\downarrow$1.32) & 5.34 ($\downarrow$0.77) \\
  & \bb{\textbf{InfoPro+MAN++ (K=4)}} & \bb{\textbf{20.16 ($\downarrow$1.53)}} & \bb{\textbf{5.17 ($\downarrow$0.94)}} \\ \hline
\multirow{4}{*}{\makecell[c]{VGG-13}} 
  & E2E & 28.41 & 9.63 \\
  & DGL~\cite{28} & 35.60 & 14.20 \\
  & DGL+MAN (K=10) & 31.99 ($\downarrow$3.61) & 10.84 ($\downarrow$3.36) \\
  & \bb{\textbf{DGL+MAN++ (K=10)}} & \bb{\textbf{30.93 ($\downarrow$4.67)}} & \bb{\textbf{10.24 ($\downarrow$3.96)}} \\ \hline
\multirow{3}{*}{\makecell[c]{ViT-Tiny}} 
  & E2E & 35.59 & 14.69 \\
  & \bb{\textbf{MAN++ (K=4)}} & \bb{\textbf{35.43 ($\downarrow$0.16)}} & \bb{\textbf{14.61 ($\downarrow$0.08)}} \\
  & \textbf{MAN++ (K=6)} & 35.51 ($\downarrow$0.08) & 14.64 ($\downarrow$0.05) \\ \hline
\multirow{3}{*}{\makecell[c]{ViT-Small}} 
  & E2E & 37.38 & 17.32 \\
  & \bb{\textbf{MAN++ (K=4)}} & \bb{\textbf{36.93 ($\downarrow$0.45)}} & \bb{\textbf{16.95 ($\downarrow$0.37)}} \\
  & \textbf{MAN++ (K=6)} & 37.15 ($\downarrow$0.23) & 17.19 ($\downarrow$0.14) \\ \hline
\multirow{3}{*}{\makecell[c]{ViT-Base}} 
  & E2E & 39.18 & 19.55 \\
  & \bb{\textbf{MAN++ (K=4)}} & \bb{\textbf{38.90 ($\downarrow$0.28)}} & \bb{\textbf{19.29 ($\downarrow$0.26)}} \\
  & \textbf{MAN++ (K=6)} & 39.03 ($\downarrow$0.15) & 19.52 ($\downarrow$0.03) \\ \hline
\end{tabular}}
\label{Table 2}
\end{table}


\begin{figure*}[htbp]
    \centering
    \includegraphics[width=0.85\textwidth]{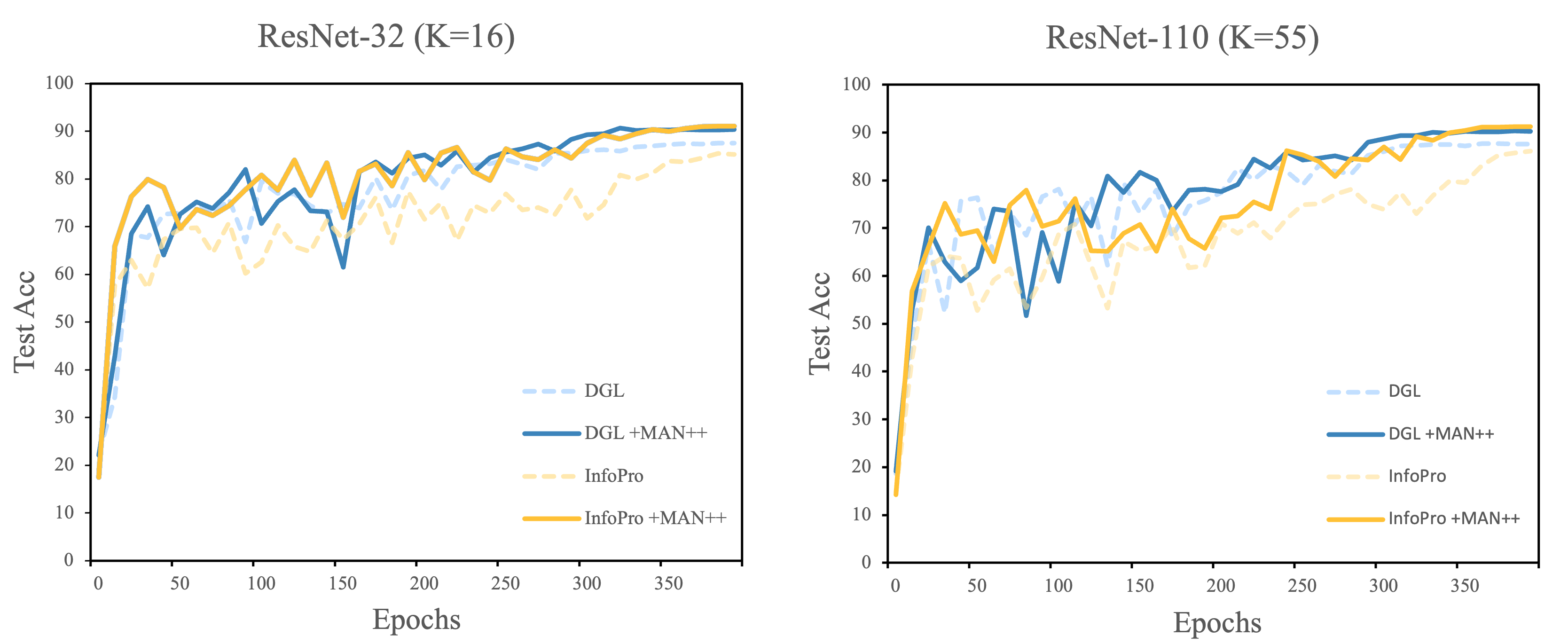}
    \caption{Training-Accuracy curves, both are utilizing the CIFAR-10 dataset.}
    \label{Figure 4}
\end{figure*}


\begin{table*}[!htbp]
  \centering
  \scriptsize
  \caption{Comparison of GPU memory usage during image-classification training.
           Values are in GB; ImageNet numbers are averaged over 8 GPUs.
           Columns with our method are shaded blue; parentheses give the relative reduction w.r.t.\ E2E (one-decimal precision).}
  \label{tab:cls_memory}
  \setlength{\tabcolsep}{6pt}
  \begin{tabular}{l l c M M}
    \toprule
    \textbf{Dataset} & \textbf{Backbone} & \textbf{E2E}
                    & \textbf{InfoPro+MAN++}
                    & \textbf{MAN++} \\ 
    \midrule
    \multirow{2}{*}{CIFAR-10}
        & ResNet-32                    &  3.37 & \textbf{(K=16) 2.45 ($\downarrow$\,27.3\%)} & -- \\
        & ResNet-110                   &  9.26 & \textbf{(K=55) 3.22 ($\downarrow$\,65.3\%)} & -- \\
    \midrule
    \multirow{6}{*}{ImageNet}
        & ResNet-101                   & 19.71 & \textbf{(K=4) 10.94 ($\downarrow$\,44.5\%)} & -- \\
        & ResNet-152                   & 26.29 & \textbf{(K=4) 14.04 ($\downarrow$\,46.6\%)} & -- \\
        & ResNeXt-101, 32$\times$8d    & 19.22 & \textbf{(K=4) 10.78 ($\downarrow$\,43.9\%)} & -- \\
        & ViT-Tiny                     &  6.69 & --                                  & \textbf{(K=6) 3.41 ($\downarrow$\,49.0\%)} \\
        & ViT-Small                    & 11.04 & --                                  & \textbf{(K=6) 5.37 ($\downarrow$\,48.7\%)} \\
        & ViT-Base                     & 22.05 & --                                  & \textbf{(K=6) 9.35 ($\downarrow$\,57.7\%)} \\
    \bottomrule
  \end{tabular}
\end{table*}

\noindent \textbf{Results on GPU memory requirement:} 
Table~\ref{tab:cls_memory} demonstrates that substituting full back-propagation with MAN++ yields substantial reductions in peak GPU demand.  
On classical CNN backbones the savings are particularly striking: ResNet-110 requires only 3.22\,GB with InfoPro+MAN++, \,$\mathbf{\downarrow 65\%}$ relative to the 9.26\,GB used by E2E training; ResNet-152 and ResNeXt-101 (32$\times$8d) each shed nearly half of their memory footprint, dropping from 26.29\,GB and 19.22\,GB to 14.04\,GB and 10.78\,GB, respectively.  
Transformer backbones benefit as well: ViT-Base runs in 9.35\,GB with MAN++, a 58\% reduction, while even the compact ViT-Tiny still saves 49\,\% of memory.  
Across all listed models the relative reduction ranges from 27\,\% (shallow ResNet-32) up to 65\,\% (deep ResNet-110), confirming that MAN++ scales favourably with depth and architectural variety.

\noindent\textbf{Accuracy retention.}  
Crucially, these memory savings incur \emph{no loss of predictive quality}.  
The same MAN++ configurations that cut the footprint by up to two-thirds also lower Top-1 / Top-5 error on ImageNet (Table~\ref{Table 2}) and reduce classification error of InfoPro method on CIFAR-10, STL-10 and SVHN (Table~\ref{Table.1}).

\subsection{Training Dynamics}
\noindent\textbf{Faster convergence.}
On ResNet-32, incorporating the MAN++ method enables our model to surpass the respective baselines within the first 40 epochs and to reach performance stability approximately 120 epochs earlier. Moreover, the time required to achieve about 90\% of the final accuracy is also significantly reduced compared to the original methods.

\noindent\textbf{Higher terminal performance.}
The final test accuracy is improved by approximately 4–5 percentage points on both ResNet-32 and ResNet-110, which verifies the effectiveness of the scalable EMA–LB mechanism in MAN++ and demonstrates its strong scalability in deeper networks.

\noindent\textbf{Stability in deep splits.}
For \(K{=}55\) the baselines plateau early and oscillate, whereas
MAN++ continues a smooth ascent before stabilising,
indicating superior optimisation stability under extreme block
partitioning.

\begin{table*}[!htbp]
  \scriptsize
  \caption{Detection results on COCO validation set with RetinaNet backbones.
           MAN++ uses $K{=}4$; E2E is the standard baseline.
           Top block: models trained from scratch for 100 epochs.
           Bottom block: models pre-trained for 100 epochs on ImageNet-1K. The GPU stands for the average memory usage per GPU.}
  \label{Table.coco}
  \centering
  \resizebox{0.7\linewidth}{!}{%
  \setlength{\tabcolsep}{2.8pt}
  \begin{tabular}{ll|cM|cM|cM|cM}
    \toprule
    \multirow{2}{*}{Setting} & \multirow{2}{*}{Backbone}
       & \multicolumn{2}{c|}{\textbf{GPU (GB)} $\downarrow$}
       & \multicolumn{2}{c|}{\textbf{mAP [\%]} $\uparrow$}
       & \multicolumn{2}{c|}{\textbf{AP$_{50}$ [\%]} $\uparrow$}
       & \multicolumn{2}{c}{\textbf{AP$_{75}$ [\%]} $\uparrow$} \\[-2pt]
    \cmidrule(lr){3-4}\cmidrule(lr){5-6}\cmidrule(lr){7-8}\cmidrule(l){9-10}
       & & E2E & \textbf{MAN++} & E2E & \textbf{MAN++} & E2E & \textbf{MAN++} & E2E & \textbf{MAN++} \\
    \midrule
    \multirow{5}{*}{Scratch}
      & ResNet-18  & 11.5 & \textbf{3.36 ($\downarrow$70.7\%)} &
                      17.2 & \textbf{18.1 ($\uparrow$0.9)} &
                      27.5 & \textbf{28.7 ($\uparrow$1.2)} &
                      18.1 & \textbf{19.1 ($\uparrow$1.0)} \\
      & ResNet-34  & 13.7 & \textbf{3.59 ($\downarrow$73.9\%)} &
                      19.3 & \textbf{19.7 ($\uparrow$0.4)} &
                      30.1 & \textbf{30.5 ($\uparrow$0.4)} &
                      \textbf{20.3} & 20.1 ($\downarrow$0.2) \\
      & ResNet-50  & 18.1 & \textbf{6.34 ($\downarrow$65.1\%)} &
                      20.1 & \textbf{20.4 ($\uparrow$0.3)} &
                      30.7 & \textbf{31.1 ($\uparrow$0.4)} &
                      19.9 & \textbf{20.5 ($\uparrow$0.6)} \\
      & ResNet-101 & 26.2 & \textbf{6.81 ($\downarrow$74.0\%)} &
                      \textbf{20.7} & 20.1 ($\downarrow$0.6) &
                      30.9 & \textbf{31.1 ($\uparrow$0.2)} &
                      20.1 & \textbf{20.8 ($\uparrow$0.7)} \\
      & ResNet-152 & 29.3 & \textbf{7.74 ($\downarrow$73.6\%)} &
                      20.5 & \textbf{20.9 ($\uparrow$0.4)} &
                      31.1 & \textbf{31.4 ($\uparrow$0.3)} &
                      20.5 & \textbf{21.3 ($\uparrow$0.8)} \\
    \midrule
    \multirow{5}{*}{Pre-train}
      & ResNet-18  & 11.5 & \textbf{3.36 ($\downarrow$70.7\%)} &
                      25.3 & \textbf{25.5 ($\uparrow$0.2)} &
                      39.5 & \textbf{40.1 ($\uparrow$0.6)} &
                      26.6 & \textbf{26.9 ($\uparrow$0.3)} \\
      & ResNet-34  & 13.7 & \textbf{3.59 ($\downarrow$73.9\%)} &
                      27.8 & \textbf{28.1 ($\uparrow$0.3)} &
                      42.2 & \textbf{42.4 ($\uparrow$0.2)} &
                      29.6 & \textbf{29.8 ($\uparrow$0.2)} \\
      & ResNet-50  & 18.1 & \textbf{6.34 ($\downarrow$65.1\%)} &
                      27.9 & \textbf{28.3 ($\uparrow$0.4)} &
                      42.8 & \textbf{43.1 ($\uparrow$0.3)} &
                      29.7 & \textbf{29.9 ($\uparrow$0.2)} \\
      & ResNet-101 & 26.2 & \textbf{6.81 ($\downarrow$74.0\%)} &
                      27.5 & \textbf{28.4 ($\uparrow$0.9)} &
                      41.7 & \textbf{43.2 ($\uparrow$1.5)} &
                      29.9 & \textbf{30.2 ($\uparrow$0.3)} \\
      & ResNet-152 & 29.3 & \textbf{7.74 ($\downarrow$73.6\%)} &
                      27.2 & \textbf{27.7 ($\uparrow$0.5)} &
                      41.1 & \textbf{42.4 ($\uparrow$1.3)} &
                      28.9 & \textbf{29.3 ($\uparrow$0.4)} \\
    \bottomrule
  \end{tabular}}
\end{table*}

\subsection{Results on Object Detection Dataset}
We conduct a comprehensive set of experiments on the COCO dataset and compared our method with the E2E training approach; the detailed results are presented in Table \ref{Table.coco}.

\noindent\textbf{Comparison with end-to-end training.}  
Across all five RetinaNet backbones in Table~\ref{Table.coco}, MAN++ reduces peak GPU memory from \,11.5–29.3\,GB to \,3.4–7.7\,GB, a relative saving of \textbf{65-74\,\%}.  
Despite this large reduction, detection accuracy is consistently better: mAP improves by $+0.2$ to $+0.9$\,pp, AP$_{50}$ by $+0.2$ to $+1.5$\,pp, and AP$_{75}$ by up to $+1.0$\,pp when compared with the standard E2E optimiser.

\noindent\textbf{Practical advantage.}  
The same $K{=}4$ configuration of MAN++ is used for every depth, yet it simultaneously lowers memory requirements and raises all three accuracy metrics.  
This favourable memory--accuracy trade-off makes MAN++ a drop-in alternative to back-propagation for large-scale object detection, enabling deeper models or larger batches on fixed hardware without retuning hyper-parameters.

\subsection{Results on Semantic Segmentation Dataset}
\noindent\textbf{Segmentation accuracy.}  
Table~\ref{tab:city_acc} shows that adding MAN++ to InfoPro (k=2) consistently outperforms both the E2E baseline and the original InfoPro (K=2) across every backbone and pre-training regime.  
Without ImageNet pre-training, MAN++ lifts mean IoU by $+1.1$–$+1.8$ pp on the lighter DeepLabV3 family and by $+1.0$–$+1.2$ pp on the stronger DeepLabV3$+$ variants, while also improving mean accuracy by up to $+1.6$ pp.  
After a 90-epoch ImageNet warm-up these gains remain: e.g.\ DeepLabV3$+$-R101 rises from 75.53\% (E2E) to 76.77\% mean IoU, and DeepLabV3-R101 reaches 74.57\%—both despite InfoPro alone harming accuracy.  
Hence, MAN++ not only recovers the accuracy lost by local training but exceeds the full back-prop baseline, delivering up to $+0.9$ pp in overall accuracy and $+1.3$ pp in mean IoU.

\noindent\textbf{GPU-memory footprint.}  
Table~\ref{tab:city_mem} confirms that the accuracy gains come at a markedly lower memory cost.  
Compared with E2E, InfoPro already cuts the peak footprint by 8–20\%, and MAN++ maintains most of this benefit: e.g.\ DeepLabV3-R101 runs in 25.4 GB versus 30.9 GB (\,\,$\downarrow$17.8\%) and DeepLabV3$+$-R50 in 25.3 GB versus 26.8 GB (\,\,$\downarrow$5.7\%).  
Crucially, MAN++ never increases memory over InfoPro by more than 1.3 pp yet converts InfoPro’s accuracy deficits into positive gains.  
Taken together, MAN++ achieves a superior memory–accuracy trade-off, raising all three evaluation metrics while preserving double-digit memory savings relative to full end-to-end training.

\begin{table*}[!htbp]
  \centering
  \scriptsize
  \caption{Results on Cityscapes. We use $K=2$,
           parentheses denote absolute change vs.\ E2E; our method columns in blue.}
  \label{tab:city_acc}
  \setlength{\tabcolsep}{4pt}
  \begin{tabular}{l l |
                  c c M|    
                  c c M|   
                  c c M}  
    \toprule
    \multirow{2}{*}{\textbf{Pretrain}} & \multirow{2}{*}{\textbf{Backbone}}&  \multicolumn{3}{c|}{\textbf{Overall Acc}}
      & \multicolumn{3}{c|}{\textbf{Mean Acc}}
      & \multicolumn{3}{c}{\textbf{Mean IoU}} \\
    \cmidrule(lr){3-5}\cmidrule(lr){6-8}\cmidrule(l){9-11}
    &
      & E2E & InfoPro & InfoPro+MAN++
      & E2E & InfoPro & InfoPro+MAN++
      & E2E & InfoPro & InfoPro+MAN++
      \\
    \midrule
    \multirow{4}{*}{Scratch}
      & DeepLabV3-R50         & 93.34 & 93.26 ($\downarrow$0.08) & \textbf{93.41 ($\uparrow$0.07)} %
                               & 68.09 & 68.02 ($\downarrow$0.07) & \textbf{69.12 ($\uparrow$1.03)} %
                               & 59.71 & 59.07 ($\downarrow$0.64) & \textbf{60.84 ($\uparrow$1.13)} \\
      & DeepLabV3-R101        & 92.67 & 92.39 ($\downarrow$0.28) & \textbf{93.05 ($\uparrow$0.38)} %
                               & 65.51 & 64.23 ($\downarrow$1.28) & \textbf{67.11 ($\uparrow$1.60)} %
                               & 56.74 & 55.23 ($\downarrow$1.51) & \textbf{58.54 ($\uparrow$1.80)} \\
      & DeepLabV3$+$-R50      & 93.06 & 92.84 ($\downarrow$0.22) & \textbf{93.18 ($\uparrow$0.12)} %
                               & 67.38 & 65.69 ($\downarrow$1.69) & \textbf{68.25 ($\uparrow$0.87)} %
                               & 58.89 & 56.58 ($\downarrow$2.31) & \textbf{60.04 ($\uparrow$1.15)} \\
      & DeepLabV3$+$-R101     & 93.14 & 92.69 ($\downarrow$0.45) & \textbf{93.21 ($\uparrow$0.07)} %
                               & 66.51 & 65.33 ($\downarrow$1.18) & \textbf{67.12 ($\uparrow$0.61)} %
                               & 58.23 & 55.91 ($\downarrow$2.32) & \textbf{59.31 ($\uparrow$1.08)} \\
    \midrule
    \multirow{4}{*}{ImageNet-1K}
      & DeepLabV3-R50         & 95.27 & 94.91 ($\downarrow$0.36) & \textbf{95.31 ($\uparrow$0.04)} %
                               & 80.83 & 80.15 ($\downarrow$0.68) & \textbf{81.24 ($\uparrow$0.41)} %
                               & 73.30 & 72.48 ($\downarrow$0.82) & \textbf{73.85 ($\uparrow$0.55)} \\
      & DeepLabV3-R101        & 95.51 & 95.19 ($\downarrow$0.32) & \textbf{95.53 ($\uparrow$0.02)} %
                               & 82.31 & 79.79 ($\downarrow$2.52) & \textbf{82.69 ($\uparrow$0.38)} %
                               & 74.41 & 70.94 ($\downarrow$3.47) & \textbf{74.57 ($\uparrow$0.16)} \\
      & DeepLabV3$+$-R50      & 95.66 & 95.41 ($\downarrow$0.25) & \textbf{95.69 ($\uparrow$0.03)} %
                               & 81.89 & 80.15 ($\downarrow$1.74) & \textbf{82.19 ($\uparrow$0.30)} %
                               & 74.61 & 72.48 ($\downarrow$2.13) & \textbf{75.34 ($\uparrow$0.73)} \\
      & DeepLabV3$+$-R101     & 95.84 & 94.93 ($\downarrow$0.91) & \textbf{95.89 ($\uparrow$0.05)} %
                               & 83.24 & 79.30 ($\downarrow$3.94) & \textbf{84.01 ($\uparrow$0.77)} %
                               & 75.53 & 71.28 ($\downarrow$4.25) & \textbf{76.77 ($\uparrow$1.24)} \\
    \bottomrule
  \end{tabular}
\end{table*}

\begin{table}[!htbp]
  \centering
  \scriptsize
  \caption{GPU memory (GB) on Cityscapes.
           Numbers in parentheses give the relative reduction with respect to E2E; 
           our method columns are shaded blue.}
  \label{tab:city_mem}
  \setlength{\tabcolsep}{5pt}
  \begin{tabular}{l c c M}
    \toprule
        \textbf{Backbone}
        & \textbf{E2E}
        & \textbf{InfoPro}
        & \textbf{InfoPro+MAN} \\
    \midrule
         DeepLabV3-R50          & 23.90  & 20.71 ($\downarrow$\,13.4\%) & \textbf{20.95 ($\downarrow$\,12.3\%)} \\
         DeepLabV3-R101         & 30.91  & 25.02 ($\downarrow$\,19.1\%) & \textbf{25.41 ($\downarrow$\,17.8\%)} \\
         DeepLabV3$+$-R50       & 26.81  & 22.67 ($\downarrow$\,\,15.5\%) & \textbf{23.29 ($\downarrow$\,\,13.2\%)} \\
         DeepLabV3$+$-R101      & 34.42  & 27.55 ($\downarrow$\,20.0\%) & \textbf{28.47 ($\downarrow$\,17.3\%)} \\
    \bottomrule
  \end{tabular}
\end{table}

\subsection{Ablation Studies}

We conduct an ablation study on the CIFAR-10 \cite{25} dataset and ImageNet dataset to assess the impact of the EMA method \cite{34}, learnable bias and the scaleable parameters in the MAN++ on performance. For this analysis, we use ResNet-32 (K=16) \cite{24} as the backbone and the original DGL \cite{28} method as a comparison baseline.

\noindent\textbf{Results on CIFAR-10.}  
Using ResNet-32 split into 16 local blocks (Table \ref{tab:ablation}\,(a)), the plain DGL baseline registers a 14.08 \% test error.  
Introducing the \emph{EMA} update alone already lowers the error to 11.07 \% (–3.01 pp), a 21 \% relative gain that confirms the benefit of propagating a smoothed inter-block signal.  
Adding a \emph{learnable bias} on top of EMA further reduces the error to 9.11 \% (–4.97 pp in total), indicating that the bias helps reconcile feature statistics across local blocks.  
Finally, equipping both EMA and bias with a \emph{scalable parameter} yields the best result, 8.75 \% (–5.33 pp, 38 \% relative), showing that the three ingredients are complementary rather than redundant.

\noindent\textbf{Results on ImageNet.}  
The same trend holds for large-scale ImageNet with ResNet-101 ($K{=}4$, Table \ref{tab:ablation}\,(b)).  
EMA alone cuts Top-1 / Top-5 error from 22.81 / 6.54 \% to 22.09 / 6.07 \% (–0.72 and –0.47 pp).  
Introducing the learnable bias amplifies the drop to 21.73 / 5.81 \% (–1.08 / –0.73 pp), and adding the scalable parameter gives the strongest performance, 21.66 / 5.72 \% (–1.15 / –0.82 pp).  
Thus, each component contributes a measurable share of the final improvement, and their combination delivers up to \textbf{5 pp} absolute error reduction on CIFAR-10 and \textbf{1.2 pp} on ImageNet.

\subsection{Qualitative Evidence for EMA–LB Complementarity}
\begin{figure*}[htbp]
    \centering
    \includegraphics[width=0.85\textwidth]{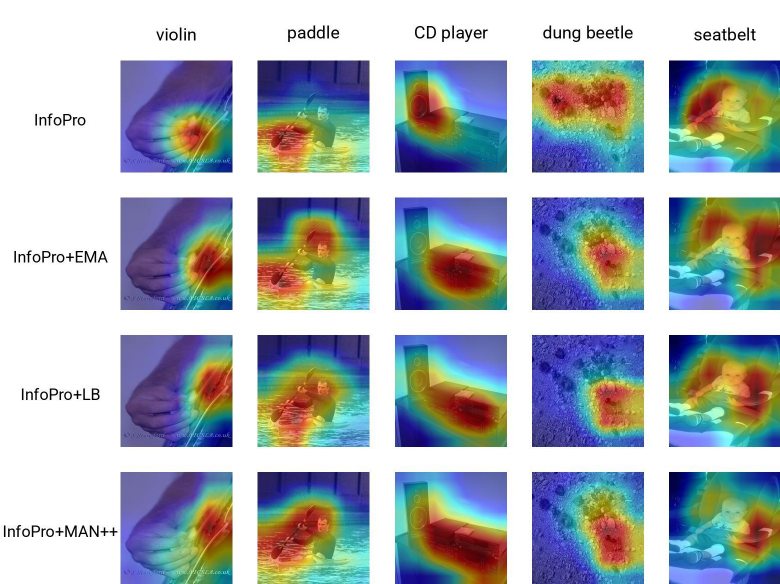}
    \caption{Grad-CAM visualizations of our methods on the ImageNet-1K validation set. All visualizations were generated using checkpoints trained for 90 epochs on the ImageNet dataset use ResNet-101 ($K=4$).}
    \label{fig:fig5}
\end{figure*}

To verify the effectiveness of our method and the complementarity of its components, we performed Grad-CAM visualizations and analyses on the ImageNet-1K validation set, as shown in Fig.~\ref{fig:fig5}.

\begin{table}[t]
  \centering
  \scriptsize
  \caption{Ablation study of MAN++ components.  
           LB denotes Learnable Bias; Scalable indicates learnable parameters are added to EMA and LB.}
  \label{tab:ablation}
  \setlength{\tabcolsep}{4pt}

  \subfloat[ResNet-32 ($K{=}16$) on CIFAR-10]{%
    \begin{tabular}{c c c c}
      \toprule
      EMA & LB & Scalable & Test Err.\,(\%) \\
      \midrule
      $\times$ & $\times$ & $\times$ & 14.08 \\
      $\checkmark$ & $\times$ & $\times$ & 11.07 ($\downarrow$\,3.01) \\
      $\checkmark$ & $\checkmark$ & $\times$ & 9.11 ($\downarrow$\,4.97) \\
      $\checkmark$ & $\checkmark$ & $\checkmark$ & \textbf{8.75 ($\downarrow$\,5.33)} \\
      \bottomrule
    \end{tabular}
  }\\[12pt]

  \subfloat[ResNet-101 ($K{=}4$) on ImageNet]{%
    \begin{tabular}{c c c c c}
      \toprule
      EMA & LB & Scalable & Top-1 Err.\,(\%) & Top-5 Err.\,(\%) \\
      \midrule
      $\times$ & $\times$ & $\times$ & 22.81 & 6.54 \\
      $\checkmark$ & $\times$ & $\times$ & 22.09 ($\downarrow$\,0.72) & 6.07 ($\downarrow$\,0.47) \\
      $\checkmark$ & $\checkmark$ & $\times$ & 21.73 ($\downarrow$\,1.08) & 5.81 ($\downarrow$\,0.73) \\
      $\checkmark$ & $\checkmark$ & $\checkmark$ & \textbf{21.66 ($\downarrow$\,1.15)} & \textbf{5.72 ($\downarrow$\,0.82)} \\
      \bottomrule
    \end{tabular}
  }
\end{table}

\noindent\textbf{Effect of EMA.}
The EMA mechanism transfers a smoothed parameter
trace from block \(j{+}1\) to block \(j\), granting each local auxiliary
classifier global context that is absent under purely myopic supervision.
As visible in the second row of Fig.~\ref{fig:fig5}, the salient
regions broaden from isolated patches to encompass the entire object such as the paddler and waterline in \textbf{canoe}, indicating richer holistic awareness.

\noindent\textbf{Effect of LB.}
The learnable bias \(b_j\) compensates for the feature-domain shift
introduced by EMA blending. The third row shows sharper,
more concentrated activations: background shelving in
\textbf{CD-player} and sand granules in \textbf{dung-beetle} are suppressed, demonstrating that LB refines the decision surface after EMA’s coarse alignment.

\noindent\textbf{Complementarity in MAN++.}
Coupling EMA and LB via the learnable scale \(s_j\) yields the best of both worlds. The fourth row combines global object coverage with precise localisation—non-object regions are strongly attenuated while key parts such as the diagonal strap and buckle in \textbf{seat-belt}) remain highlighted. The Grad-CAM visualisations confirming that EMA supplies essential inter-block context, LB resolves feature
mis-alignment, and their synergy in \textbf{MAN++} produces gradients that are simultaneously comprehensive and precise.

\subsection{Representation Similarity Analysis}

We further evaluate the quality of the learned representations with Centered Kernel Alignment (CKA)~\cite{kornblith2019similarity}.
CKA measures how closely the features of a given method align with those produced by an end‑to‑end (E2E) baseline; a score of 1 indicates identical representations.

\begin{figure*}[t]
    \centering
    \includegraphics[width=0.95\textwidth]{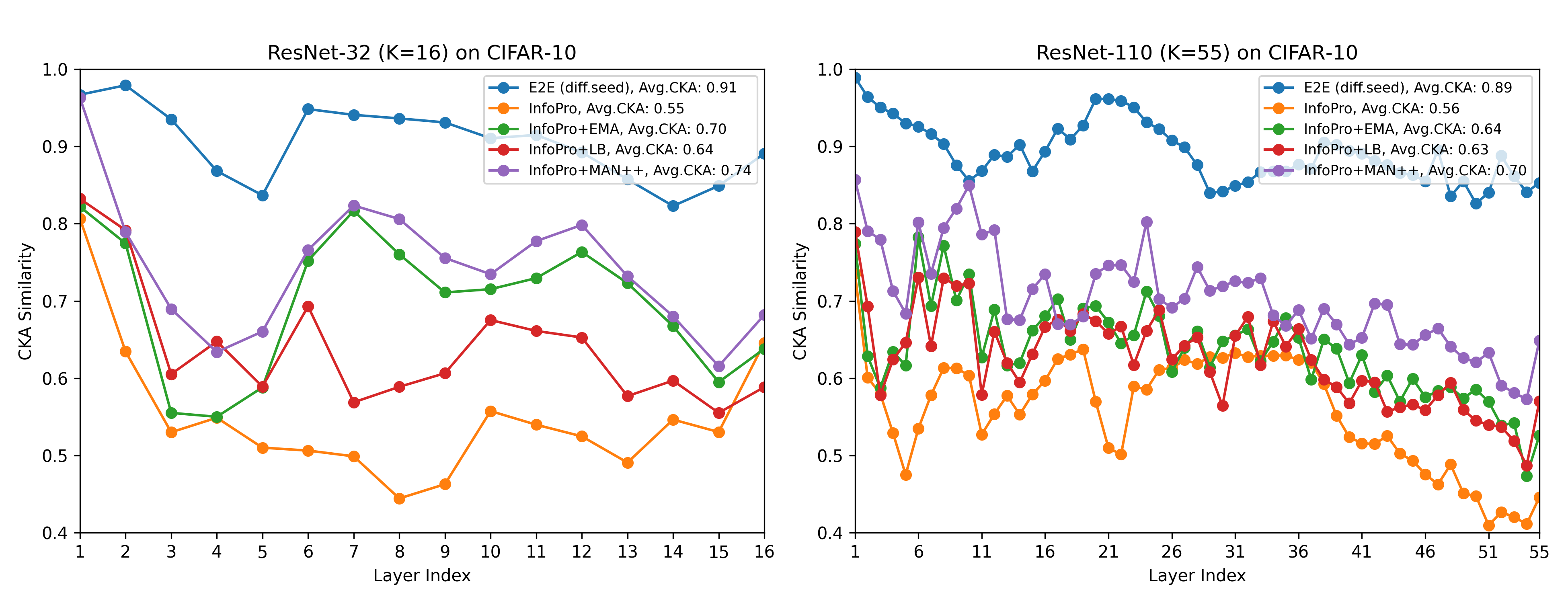}
    \caption{Assessment of Similarity in Layer-wise Representations. We use Centered Kernel Alignment (CKA) {\cite{kornblith2019similarity}} to quantify the degree of similarity in the layer-wise representations between the E2E backpropagation and our proposed MAN++.}
    \label{fig:cka_all}
\end{figure*}

Fig.\ref{fig:cka_all} reports layer‑wise CKA on CIFAR‑10 for ResNet‑32 and ResNet‑110.
Several observations stand out:
\begin{itemize}
    \item \textbf{E2E training forms the upper bound.}
    The blue curves remain highest across all layers, confirming that full back‑propagation still learns the most consistent global features.
    \item \textbf{InfoPro shows the largest deviation.}
    Without any extra auxiliary mechanism, the orange curves drop below 0.60 in early layers and never recover to the E2E envelope, illustrating the representation drift induced by purely local objectives.
    \item \textbf{Adding either EMA or a learnable bias narrows the gap.} Green (+EMA) and red (+LB) curves raise the average CKA by 8–15 points relative to InfoPro, with EMA delivering the stronger early‑layer improvement—evidence that momentum feedback mitigates the locality bottleneck.
    \item \textbf{MAN++ achieves the best and most stable alignment.} The purple curves not only attain the highest average scores among local methods (0.74 on ResNet‑32 and 0.70 on ResNet‑110) but also exhibit the largest gains in the first two and last three blocks.
    Coupled with our linear‑separability results, this suggests that MAN++ steers the early blocks to learn more general, task‑oriented features, while simultaneously enhancing the semantic consistency of the deepest layers.
    \item \textbf{Depth amplifies the challenge, yet the trend persists.}
    On the 55‑layer network the absolute gap to E2E widens, but MAN++ still outperforms all other local variants, confirming its scalability.
\end{itemize}
Overall, the analysis corroborates that MAN++ enables effective information exchange across gradient‑isolated blocks, alleviating the myopia problem that plagues existing supervised local‑learning methods.

\subsection{Training speed analysis of MAN++}

Compared to the E2E approach, Local Learning methods introduce a substantial number of auxiliary network parameters. As a result, although they offer significant advantages in terms of GPU memory consumption, training speed has remained a major bottleneck. Recently, research in the Local Learning field has led to the development of a multi-GPU accelerated training strategy, PPLL~\cite{ppll}. Unlike traditional DataParallel (DP) or DistributedDataParallel (DDP) methods~\cite{ddp1}, PPLL assigns the $K$ network blocks obtained from local learning to $K$ different GPUs—placing one local block and its corresponding auxiliary network on each GPU. A message queue is used to store the latest outputs from each block, enabling fully parallel multi-GPU training and greatly reducing communication costs. We adopted the PPLL method to benchmark training time for multi-GPU setups, and compared the results with single-GPU and multi-GPU DDP training times. The pseudocode for our multi-GPU PPLL experiments is shown in Algorithm~\ref{alg:ppll_manpp}.

\textbf{Settings.} To ensure the credibility of our comparisons, we set the batch size to 128 for single GPU experiments on ImageNet, and to 16 for both COCO and CityScapes. When using DDP, the batch size is multiplied by the number of GPUs. For PPLL, the batch size is kept the same as in the single GPU setting.

The detailed experimental results are presented in Table~\ref{tab:speed}. Our analysis leads to the following main conclusions:

\noindent \textbf{Single--GPU overhead.} Adding MAN++ increases per-epoch time by $20$-$30\%$ (e.g.\ ResNet-101: $2242\!\to\!2726$\,s), because each forward step now includes auxiliary heads and EMA updates.

\noindent \textbf{Limited benefit of naive DDP.} Replicating the whole network via DistributedDataParallel
shortens epochs, yet MAN+++ DDP remains $14$-$20\%$ slower than E2E, since every GPU still holds every auxiliary head, incurring redundant compute and gradient synchronisation.

\noindent \textbf{PPLL turns the tables.} Mapping one local block~$L_i$ and its head~$H_i$ to one GPU removes head replication and virtually all inter GPU gradient traffic. Consequently MAN++ with PPLL is consistently faster than DDP and usually on par with or even faster than E2E training; see ViT-Tiny (\,636$\!\to\!$458\,s) or ResNet-152 on COCO (\,4346$\!\to\!$3328\,s).

\noindent The absolute saving scales with model capacity
(ViT-B/16 gains $\approx$725\,s/epoch), while even a two segment setup (Cityscapes) still secures a $\sim$15\,\% boost. Hence PPLL eliminates the chief speed bottleneck of MAN++ without sacrificing its memory advantage.

\begin{algorithm}[ht]
\caption{PPLL for MAN++ when GPUs $=K$}
\label{alg:ppll_manpp}
\begin{algorithmic}[1]
\Require dataset $\mathcal{D}$, epochs $E$, local blocks $L_0,\dots,L_{K-1}$ with auxiliary heads $H_0,\dots,H_{K-2}$

\State \textbf{Initialisation}
\State map block $L_i$ \textbf{and} head $H_i$ (if $i<K-1$) to GPU $i$
\State create queues $\texttt{inQ}[i]$ and $\texttt{outQ}[i]$ for all $i$
\State push a dummy tensor into $\texttt{inQ}[0]$ \Comment{prime the pipeline}

\Procedure{Worker}{$i$} \Comment{executes on GPU $i$}
    \While{training not finished}
        \State $x \gets \texttt{inQ}[i].\mathrm{pop}()$
        \State $x \gets L_i(x)$                                          \Comment{forward}
        \If{$i < K-1$}                                                   \Comment{no head on the last block}
            \State $y \gets H_i(x)$
            \State $\ell \gets \mathcal{L}(y,\text{label})$
            \State \Call{Backward}{$\ell$}                               \Comment{local backward}
            \State \Call{EMA\_update}{$H_i,L_{i+1},s_i$}
            \State $x \gets \mathrm{detach}(x)$
        \EndIf
        \If{$i < K-1$}
            \State $\texttt{outQ}[i].\mathrm{push}(x)$
        \EndIf
        \If{$i = 0$}                                                     \Comment{first GPU feeds new data}
            \State $(x',\_) \gets \mathcal{D}.\mathrm{next\_batch}()$
            \State $\texttt{inQ}[0].\mathrm{push}(x')$
        \EndIf
        \If{$i < K-1$}
            \State $\texttt{inQ}[i+1] \gets \texttt{outQ}[i]$
        \EndIf
    \EndWhile
\EndProcedure

\For{$e\gets1$ \textbf{to} $E$}
    \State wait until all \textsc{Worker}s finish epoch $e$
    \State synchronise BN statistics of block $L_{K-1}$ (optional)
    \State update the learning rate scheduler
\EndFor

\end{algorithmic}
\end{algorithm}

\section{Theoretical Analysis}

\subsection{Notation for Theoretical Analysis}
In this section, unless otherwise stated, $L$ denotes the number of backbone layers and the backbone is evenly split into $K$ local blocks, each with length $B=\frac{L}{K}$.
We use $b$ (or $j$) to index local blocks.
Layer-wise parameter size is $p_i=\lVert W_i\rVert_0$ and $\bar p=\frac{1}{L}\sum_{i=1}^L p_i$.
We denote the layer-size imbalance ratio by $\rho_p=\frac{p_{\max}}{p_{\min}}$.
The relative size of the learnable bias is controlled by $\beta\ll 1$ (and $\beta_F,\beta_A$ for FLOPs/activation overhead, respectively).
In the FLOPs analysis, $\varepsilon$ denotes the target relative FLOPs overhead; in the memory analysis, $\rho_M$ denotes the target memory ratio.
In the convergence analysis, $L_s$ denotes the smoothness constant (to avoid confusion with the layer count $L$).

\subsection{Parameter Complexity}
\paragraph{Setup}
Assume a backbone of $L$ layers with individual sizes
$p_i=\lVert W_i\rVert_0$ and let
\(p_{\min}\le p_i\le p_{\max}\)
with imbalance ratio $\rho_p=\frac{p_{\max}}{p_{\min}}\ge1$.
The network is evenly split into
$K$ local blocks of length $B=\frac{L}{K}$,
$S_b=\{(b-1)B+1,\dots,bB\}$.
MAN++ attaches an auxiliary head to the first $K-1$ blocks,
formed by the \emph{first} layer of the successive block and a
learnable bias~$b_b$;
we set $\lVert b_b\rVert_0=\beta\,
      \bigl\lVert\tilde W_{b+1}^{(1)}\bigr\rVert_0$
with $\beta\ll1$.

\paragraph{Extra parameters}
Let $\bar p=\frac1L\sum_{i=1}^{L}p_i$ be the mean layer size.
The \emph{expected} overhead of MAN++ is
\begin{equation}
  \mathbb{E}[\Delta P]=(K-1)(1+\beta)\,\bar p .
\end{equation}
Since every $\tilde W_{b+1}^{(1)}$ contains between $p_{\min}$ and
$p_{\max}$ parameters, the overhead admits the bounds
\begin{equation}
  (1+\beta)(K-1)p_{\min}
  \;\le\;
  \Delta P
  \;\le\;
  (1+\beta)(K-1)p_{\max}
\end{equation}

\paragraph{Relative size}
The E2E baseline satisfies
$Lp_{\min}\le P_{\mathrm{E2E}}\le Lp_{\max}$.
Consequently,
\begin{equation}
\label{eq:param_bounds}
  1+\frac{1+\beta}{B}\!\Bigl(1-\frac1K\Bigr)
  \;\le\;
  \frac{P_{\mathrm{MAN++}}}{P_{\mathrm{E2E}}}
  \;\le\;
  1+\rho_p\,\frac{1+\beta}{B}\!\Bigl(1-\frac1K\Bigr)
\end{equation}

\paragraph{Discussion}
\begin{itemize}
\item For deep backbones ($B\gg1$) both bounds converge to~$1$,
      hence the overhead is negligible.
\item For extreme splitting ($K\!\to\!L$, $B\!\to\!1$) the worst‑case growth is
      $\approx1+\rho_p(1+\beta)$; under mild imbalance
      ($\rho_p\!\approx\!1$) and tiny bias ($\beta\!\ll\!1$) the total
      parameter count remains below $2P_{\mathrm{E2E}}$.
\end{itemize}

\begin{table*}[!htbp]
  \centering
  \scriptsize
  \caption{The training speed across different datasets and backbones is reported in the table. 
           To ensure reliability, all experiments were conducted on Nvidia~A100\,GPUs, 
           and the times shown represent the \textbf{average training time per epoch} 
           measured after 30~epochs of training.}
  \label{tab:speed}
  \setlength{\tabcolsep}{5pt}
  \renewcommand{\arraystretch}{1.15}
  \begin{tabular}{l l l c c c}
    \toprule
    \textbf{Dataset} & \textbf{Backbone} & \textbf{Method} &
    \textbf{Single\,GPU} & \textbf{$K$\,GPUs (DDP)} & \textbf{$K$\,GPUs (PPLL)} \\
    \midrule

    \multirow{8}{*}{ImageNet}
        & \multirow{2}{*}{ResNet‑101}
          & E2E                         & 2242s  &  795s &  --- \\
        &  & \bb{InfoPro+MAN++ ($K{=}4$)}    & \bb{2726s}  &  \bb{973s} &  \bb{726s} \\
    \cmidrule(lr){2-6}
        & \multirow{2}{*}{ResNet‑152}
          & E2E                         & 3299s  & 1135s &  --- \\
        &  & \bb{InfoPro+MAN++ ($K{=}4$)}    & \bb{3758s}  & \bb{1298s} & \bb{1018s} \\
    \cmidrule(lr){2-6}
        & \multirow{2}{*}{ViT‑Tiny/16}
          & E2E                         & 1282s  &  503s &  --- \\
        &  & \bb{MAN++ ($K{=}4$)}            & \bb{1613s}  &  \bb{636s} &  \bb{458s} \\
    \cmidrule(lr){2-6}
        & \multirow{2}{*}{ViT‑Base/16}
          & E2E                         & 9103s  & 2976s &  --- \\
        &  & \bb{MAN++ ($K{=}4$)}            & \bb{11433s} & \bb{3620s} & \bb{2895s} \\
    \midrule

    \multirow{6}{*}{COCO}
        & \multirow{2}{*}{ResNet‑34}
          & E2E                         &  9407s & 3137s &  --- \\
        &  & \bb{MAN++ ($K{=}4$)}            & \bb{10903s} & \bb{3667s} & \bb{2869s} \\
    \cmidrule(lr){2-6}
    
    & \multirow{2}{*}{ResNet‑101}
          & E2E                         &  10654s & 3481s &  --- \\
        &  & \bb{MAN++ ($K{=}4$)}            & \bb{12271s} & \bb{4081s} & \bb{3257s} \\
    \cmidrule(lr){2-6}
    
        & \multirow{2}{*}{ResNet‑152}
          & E2E                         & 11076s & 3758s &  --- \\
        &  & \bb{MAN++ ($K{=}4$)}            & \bb{12689s} & \bb{4346s} & \bb{3328s} \\
    \midrule

    \multirow{4}{*}{City\-Scapes}
        & \multirow{2}{*}{DeepLabV3‑R50}
          & E2E                         &   80s  &   52s &  --- \\
        &  & \bb{InfoPro+MAN++ ($K{=}2$)}    &  \bb{103s}  &   \bb{64s} &   \bb{54s} \\
    \cmidrule(lr){2-6}
        & \multirow{2}{*}{DeepLabV3‑R101}
          & E2E                         &   95s  &   58s &  --- \\
        &  & \bb{InfoPro+MAN++ ($K{=}2$)}    &  \bb{131s}  &   \bb{79s} &   \bb{61s} \\
    \bottomrule
  \end{tabular}
\end{table*}

\subsection{FLOPs analysis}
Let the network contain $L$ layers that are evenly split into $K$ local
blocks.  We assume a homogeneous per–layer cost of $F$ FLOPs for one
\emph{forward} or \emph{backward} pass.

\subsubsection{E2E baseline}
Each update requires one forward \emph{and} one backward sweep:
\begin{equation}
\mathrm{FLOPs}_{\mathrm{E2E}} \;=\; 2 L F
\end{equation}

\subsubsection{MAN++}
In addition to the backbone cost $2LF$, the first $K{-}1$ blocks own an
auxiliary head consisting of
(i)~a copy of the \emph{first} layer of the succeeding block
and (ii)~a learnable bias~(LB).  
The copy layer contributes $2F$ FLOPs
(forward\,+\,backward) whereas the LB adds
$\beta_F F$ FLOPs, where $\beta_F\!\ll\!1$
(e.g.\ $\beta_F\!\approx\!0.02$ for a single channel–wise scale).

Hence the total update cost is
\begin{equation}
\mathrm{FLOPs}_{\mathrm{MAN++}}
= 2LF + (K-1)\,(2+\beta_F)F
\end{equation}

\subsubsection{Relative overhead}
\begin{equation}
\label{eq:flop_ratio}
\frac{\mathrm{FLOPs}_{\mathrm{MAN++}}}{\mathrm{FLOPs}_{\mathrm{E2E}}}
= 1 + \frac{K-1}{L}\Bigl(1+\tfrac{\beta_F}{2}\Bigr)
\end{equation}

\noindent
\textbf{Edge cases.}
(i)~If $K=L$ (one layer per block) the factor
approaches $2+\frac{\beta_F}{2}$, i.e.\ a worst‑case
increase of $\approx$\,101\,\% when $\beta_F=0.02$.
(ii)~For the common regime $K\!\ll\!L$ the term $\frac{(K{-}1)}{L}$ is small
and the overhead falls below~10\,\%.

\subsubsection{Practical guideline}
Given a desired overhead $\varepsilon$, Eq.\,\eqref{eq:flop_ratio}
implies
\begin{equation}
\label{eq.14}
K \;\le\;
1 + \frac{\varepsilon L}{1+\frac{\beta_F}{2}}
\end{equation}
so for $\varepsilon=0.1$ and $\beta_F=0.02$
networks as deep as $L=101$ can be safely
partitioned into up to $K=11$ blocks while keeping the FLOPs increase
within 10\,\%.

\noindent\textbf{Choosing the number of local blocks $K$ in practice.}
Eq.~\ref{eq.14} provides a practical upper bound on $K$ under a target FLOPs overhead $\epsilon$, where $L$ denotes the network depth under a chosen counting convention.
While using the canonical depth (e.g., $L{=}50/101$ for ResNet-50/101) is theoretically valid, it often leads to overly large $K$ and thus significantly increases the wall-clock training time due to excessive pipeline stages.
In practice, we therefore choose a coarser partition granularity for ResNets by splitting at the level of \texttt{Bottleneck} units, and denote the corresponding depth by $L_b$.
Under this convention, the maximum $L_b$ is $17$ for ResNet-50 and $34$ for ResNet-101.
With a 10\% FLOPs budget, the auxiliary overhead is typically kept within \(\approx 2\) bottleneck units for ResNet-50 and \(\approx 3\) units for ResNet-101, leading to typical choices of $K{=}3$ and $K{=}4$, respectively.

\subsection{GPU Memory analysis}
Let $L$ be the total number of layers and let the
network be split into $K$ equal blocks,
each containing $B=\frac{L}{K}$ backbone layers.
Denote by $A$ the activation footprint of a
single layer for one mini‑batch and by
$\beta_A\!\ll\!1$ the relative activation cost of the
learnable bias~(LB).

\subsubsection{E2E baseline.}
Storing all forward activations for back‑propagation yields
\begin{equation}
M_{\text{E2E}} = L\,A
\end{equation}

\subsubsection{MAN++ on a single GPU.}
During the local update of block~$j$  only (i) its $B$ backbone layers
and (ii) its auxiliary head are resident in memory.
The head consists of one layer copy and one LB, hence
\begin{equation}
M_{\text{blk}} = BA \;+\; (1+\beta_A)A
\end{equation}
Since blocks are processed sequentially, the global peak is
$M_{\text{MAN++}}=M_{\text{blk}}$ and
\begin{equation}
\frac{M_{\text{MAN++}}}{M_{\text{E2E}}}
  = \frac{BA + (1+\beta_A)A}{LA}
  = \frac{1}{K} + \frac{1+\beta_A}{L}
\label{eq:mem_ratio}
\end{equation}

\subsubsection{MAN++ with PPLL (one block per GPU).}
Each GPU now holds \emph{its} block only, hence the same peak
$M_{\text{blk}}$ applies, while an E2E–DDP replica would still keep
$LA$ activations.  Therefore Eq.\,\eqref{eq:mem_ratio} also quantifies
the memory advantage in the multi‑GPU regime.

Given a desired memory ratio $\rho<1$, Eq.\,\eqref{eq:mem_ratio}
implies the lower bound
\begin{equation}
K \;\ge\; \frac{1}{\rho - \tfrac{1+\beta_A}{L}},
\quad\text{provided}\;
\rho>\frac{1+\beta_A}{L}
\end{equation}
For example, with $L=50$, $\beta_A=0.02$ and a target
$\rho=0.25$, splitting into $K\!\ge\!4$ blocks suffices to keep the
peak activation memory below 25\,\% of the E2E baseline.

\subsection{Convergence of block–wise SGD}
Assume each local surrogate loss $\tilde F_j$ is $L_s$–smooth and its
stochastic gradient has bounded variance $\sigma^2$.
Let $\theta_t$ be the parameter after $t$ updates and denote by
$\alpha\in(0,1)$ the EMA decay and by $s_j\in(0,1]$ the learnable
scaling in MAN\texttt{++}.
Following~\cite{gidel2018variational}, the bias between the true and
local gradients satisfies
\begin{equation}
\bigl\|\nabla\tilde F_j(\theta_t)-\nabla F(\theta_t)\bigr\|
       \;\le\; c_1\,(1-s_j)\,\alpha^{B}
\end{equation}
where $c_1$ depends on inter‑block Lipschitz constants.
Running sequential block‑wise SGD with stepsizes
$\eta_t\!\propto\!\frac{1}{\sqrt{t}}$ yields
\begin{equation}\label{eq:conv_sgd}
\min_{0\le t<T}\!
      \mathbb{E}\bigl\|\nabla F(\theta_t)\bigr\|^2
      \;=\;
      \mathcal{O}\!\Bigl(\tfrac{1}{\sqrt{B\,T}}\Bigr)
      \;+\;
      \mathcal{O}\!\Bigl(\tfrac{1+\beta}{K}\Bigr)
      \;+\;
      \mathcal{O}\!\Bigl(\tfrac{c_1}{T}\Bigr)
\end{equation}
The first term is the usual stochastic noise, the second stems from
the auxiliary heads ($\beta\!\ll\!1$ for LB), while the third decays
with a faster EMA (smaller $\alpha$) or larger $s_j$.
When $K\!\ll\!L$ the $\frac{1}{K}$ factor dominates, explaining the empirical
speed–accuracy trade‑off.

\subsection{Optimality gap to full back‑propagation}
Let $\theta^\star$ be a global minimiser of $F$ and
$\tilde\theta^\star$ the optimum attained by MAN++.
Under $L_s$–smoothness and the Polyak–Łojasiewicz condition,
\begin{equation}
\begin{aligned}
F(\tilde\theta^\star) - F(\theta^\star)
&\;\le\;
C\!\Bigl[
      \frac{1+\beta}{K}
      \;+\;
      c_1\,(1-\bar s)\,\alpha^{B}
   \Bigr],
\\[4pt]
\bar s
&= \frac{1}{K-1}\sum_{j=1}^{K-1}s_j
\end{aligned}
\end{equation}
Hence the gap vanishes at rate $\mathcal{O}(\frac{1}{K})$ provided
$\bar s\!\to\!1$ and $\alpha<1$, corroborating the empirical trend
that deeper splits benefit more from MAN++.

\subsection{Extension to Adam}
With Adam ($\beta_1,\beta_2\!<\!1$) and
stepsizes $\eta_t=\frac{\eta}{\sqrt{t}}$,  
\cite{reddi2019convergence} show that under bounded gradients
the adaptive update enjoys
\(
\sum_{t=1}^{T}\frac{\eta_t}{\sqrt{v_t}}\ge c\sqrt{T}.
\)
Combining this with the gradient–bias bound yields
\begin{equation}
\begin{aligned}
  \min_{0\le t<T}\,
  \mathbb{E}\!\bigl\|\nabla F(\theta_t)\bigr\|^2
  &\;\le\;
  \mathcal{O}\!\Bigl(
      \tfrac{\sqrt{\log T}}{\sqrt{B\,T}}
  \Bigr)
\\
  &\quad+\,
  \mathcal{O}\!\Bigl(\tfrac{1+\beta}{K}\Bigr)
\\
  &\quad+\,
  \mathcal{O}\!\Bigl(c_1(1-\bar s)\alpha^{B}\Bigr).
\end{aligned}
\end{equation}
Thus MAN++ preserves Adam’s
$\mathcal{O}(\sqrt{\frac{\log T}{\!T}})$ convergence while adding the same
\(\tfrac{1+\beta}{K}\) and EMA bias terms as in
Eq.\,\eqref{eq:conv_sgd}.

\subsection{Generalization Error Bound}

\paragraph{Assumptions}
Let the empirical loss be
$\hat F_n(\theta)=\frac1n\sum_{i=1}^n\ell(f_\theta(x_i),y_i)$
with a $1$‑Lipschitz bounded loss
$\ell\!:\mathbb{R}\times\mathcal{Y}\!\to\![0,1]$.
Denote by $\Theta_{\mathrm{E2E}}$ the hypothesis class of the
end‑to‑end baseline and by $\Theta_K$ the class after splitting into
$K$ local blocks augmented by MAN++ heads.
For simplicity every backbone block is $\kappa_j$‑Lipschitz and the extra
layer$+$bias of MAN++ own constants
$\lambda$ and $\lambda_b\ll\lambda$.

\paragraph{Rademacher complexity}
Using the standard composition result
$\mathfrak{R}_n(f\!\circ\!g)\le L_f\,\mathfrak{R}_n(g)$,
iteration over $K$ blocks yields
\begin{equation}
\mathfrak{R}_n(\Theta_K)
\;\le\;
\Bigl(\prod_{j=1}^{K-1}\!\kappa_j\Bigr)\,
(1+\lambda+\lambda_b)\,
\mathfrak{R}_n(\Theta_{\mathrm{E2E}})
\end{equation}

\paragraph{Generalization gap}
With probability at least $1-\delta$,
\begin{equation}
\label{eq:gen_gap}
\begin{aligned}
  F(\theta) - \hat F_n(\theta)
  \;\le\;&\;
  2\,
  \Bigl(\textstyle\prod_{j=1}^{K-1}\kappa_j\Bigr)\,
  (1+\lambda+\lambda_b)\,
  \mathfrak{R}_n(\Theta_{\mathrm{E2E}})
\\[4pt]
  &\;+\;
  \mathcal{O}\!\Bigl(
      \sqrt{\tfrac{\log(\frac{1}{\delta})}{n}}
    \Bigr)
\end{aligned}
\end{equation}

\paragraph{Implication}
If batch normalization or spectral clipping keeps all
$\kappa_j\!\le\!c<1$ and the auxiliary bias is lightweight
($\lambda_b\!\ll\!1$), then
\begin{equation}
  \Delta_{\mathrm{gen}}
  =\mathcal{O}\!\bigl(c^{K-1}(1+\lambda+\lambda_b)\bigr)
  \xrightarrow[K\;\text{large}]{} 0
\end{equation}
so MAN++ does not increase the generalization gap
compared with the E2E baseline.
\section{Conclusion}
In this paper, we propose the Momentum Auxiliary Network++ (MAN++) method, which addresses the short-sightedness problem that commonly arises in the early optimization stages of supervised local learning. By facilitating information exchange between gradient-isolated local modules, MAN++ substantially narrows the performance gap between supervised local learning methods and end-to-end approaches in deep learning. We conduct extensive experiments on image classification, object detection, and semantic segmentation tasks using various architectures, including ResNets and ViTs. The results demonstrate the effectiveness of our method, showing that MAN++ achieves competitive performance compared to E2E training while significantly reducing GPU memory consumption. Furthermore, we explore the training speed of local learning on multi-GPU systems. Leveraging the PPLL approach, we show that local learning not only conserves memory without sacrificing accuracy, but also achieves training speeds comparable to those of end-to-end methods.

\bibliographystyle{IEEEtran}
\bibliography{IEEEabrv,ref}


\mybio{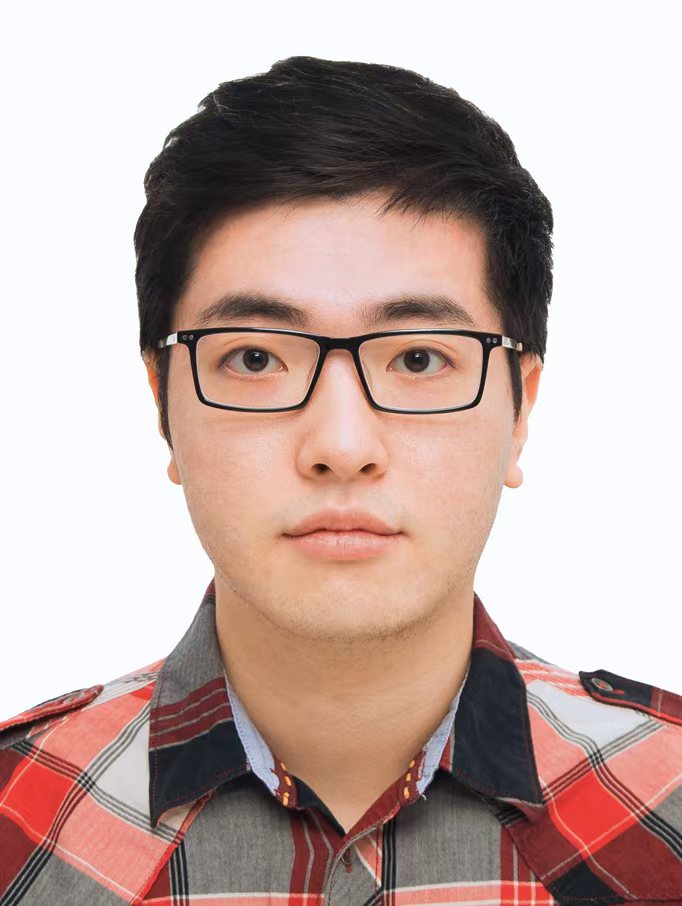}{Junhao Su}{is currrenly an algorithm engineer at Meituan. He recieved his M.S. degree in Southeast University. He has published several papers at conferences such as ECCV and AAAI. His research interests mainly include vision understanding, large language model, efficient training method, and vision agent.}

\vspace{15mm}

\mybio{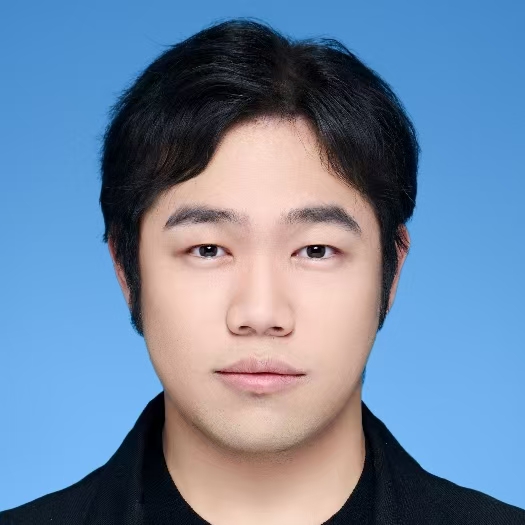}{Feiyu Zhu}{graduated from the School of Optical Engineering and Computer Technology, University of Shanghai for Science and Technology with a master's degree in 2024. He is currently working in the direction of artificial intelligence at Attrsense. His current research interests include computer vision and model lightweighting.}

\vspace{15mm}

\mybio{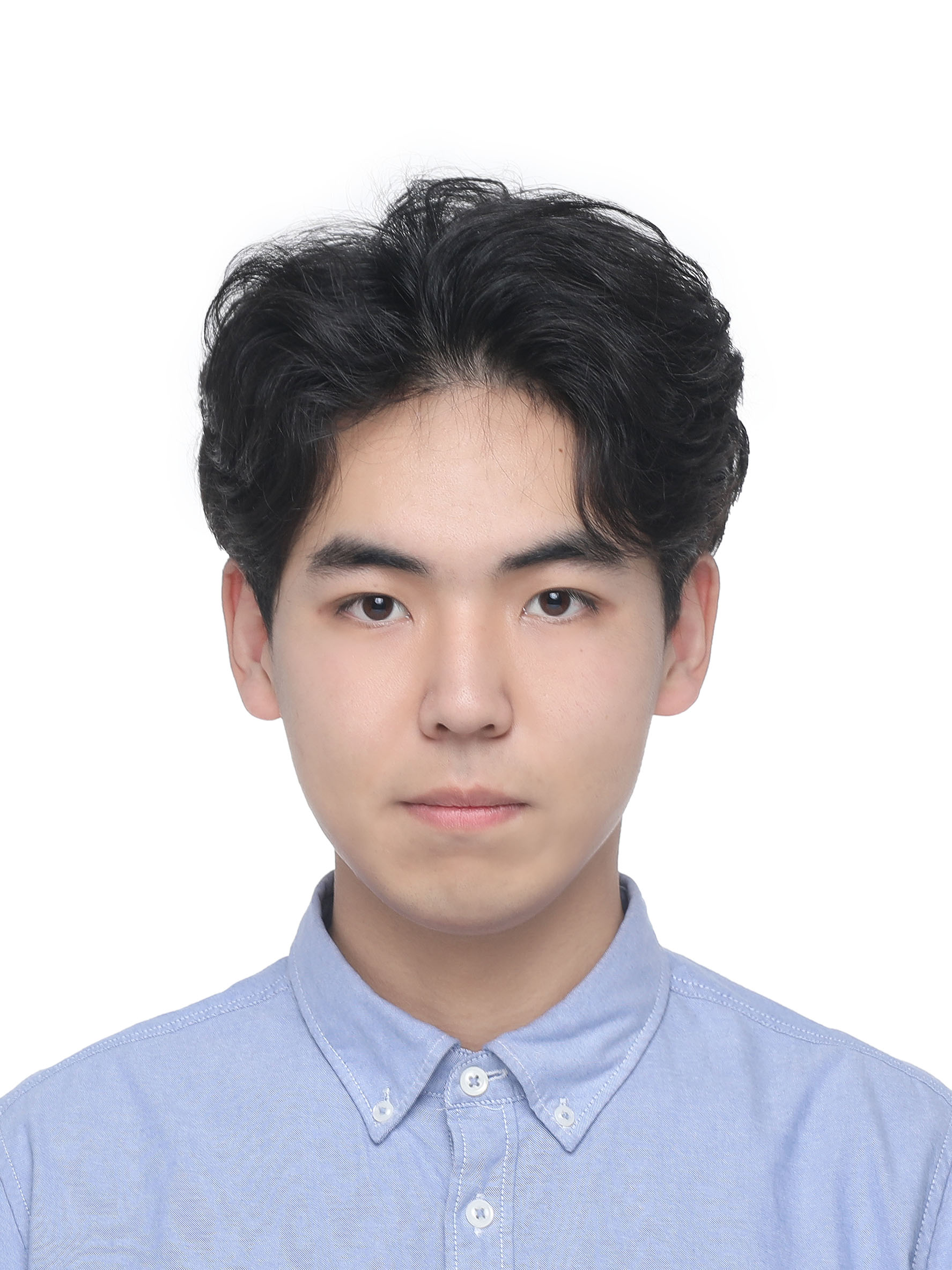}{Hengyu Shi}{is currently serves as an Algorithm Engineer at Meituan, specializing in image generation and understanding. He graduated from the University of Science and Technology Beijing.}

\vspace{15mm}
\mybio{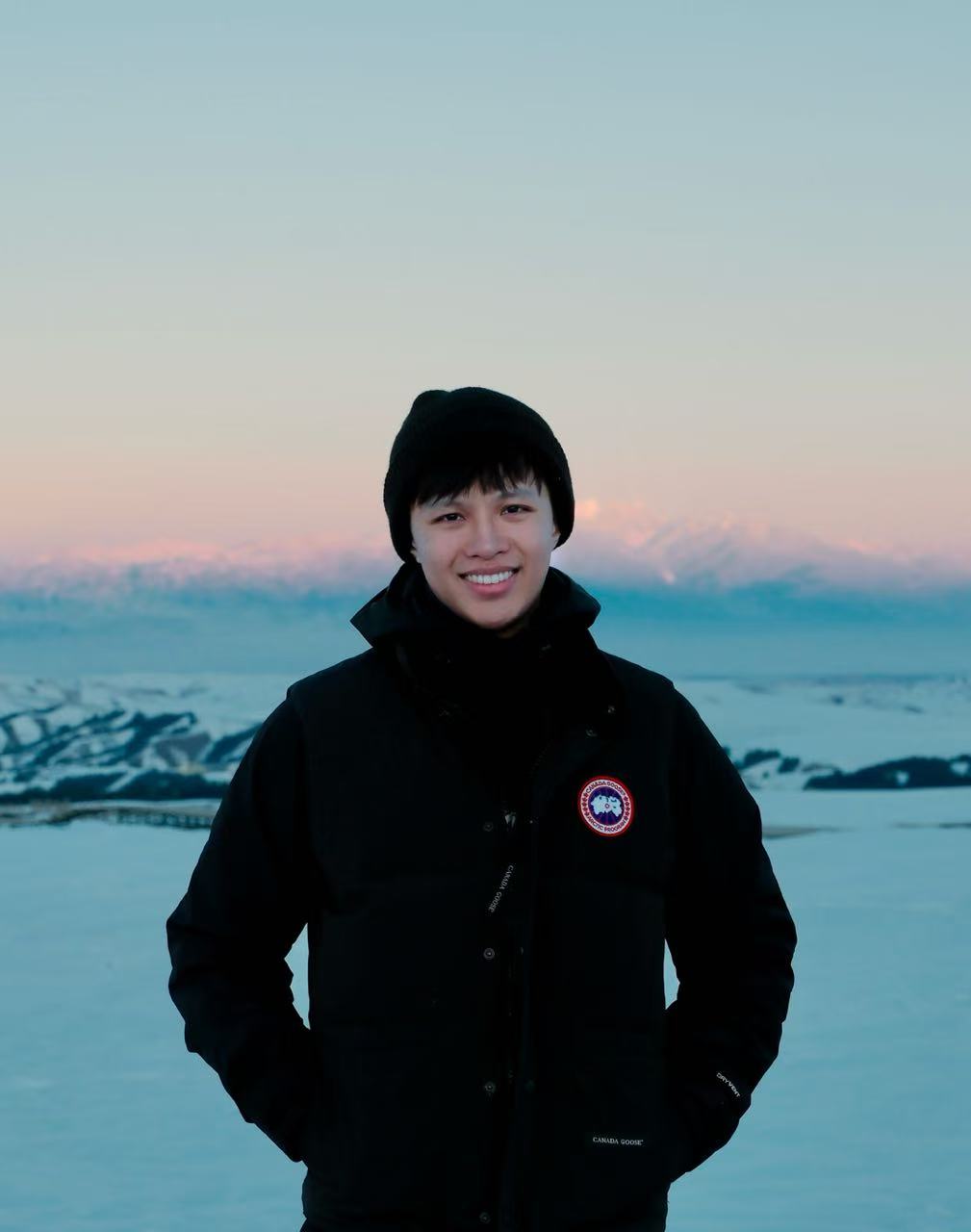}{Tianyang Han}{is currrenly an algorithm engineer at Meituan. He recieved his M.S. degree in  Hong Kong Polytechnic University. His research interests include multi-modal large language model and image generation.}

\vspace{15mm}
\mybio{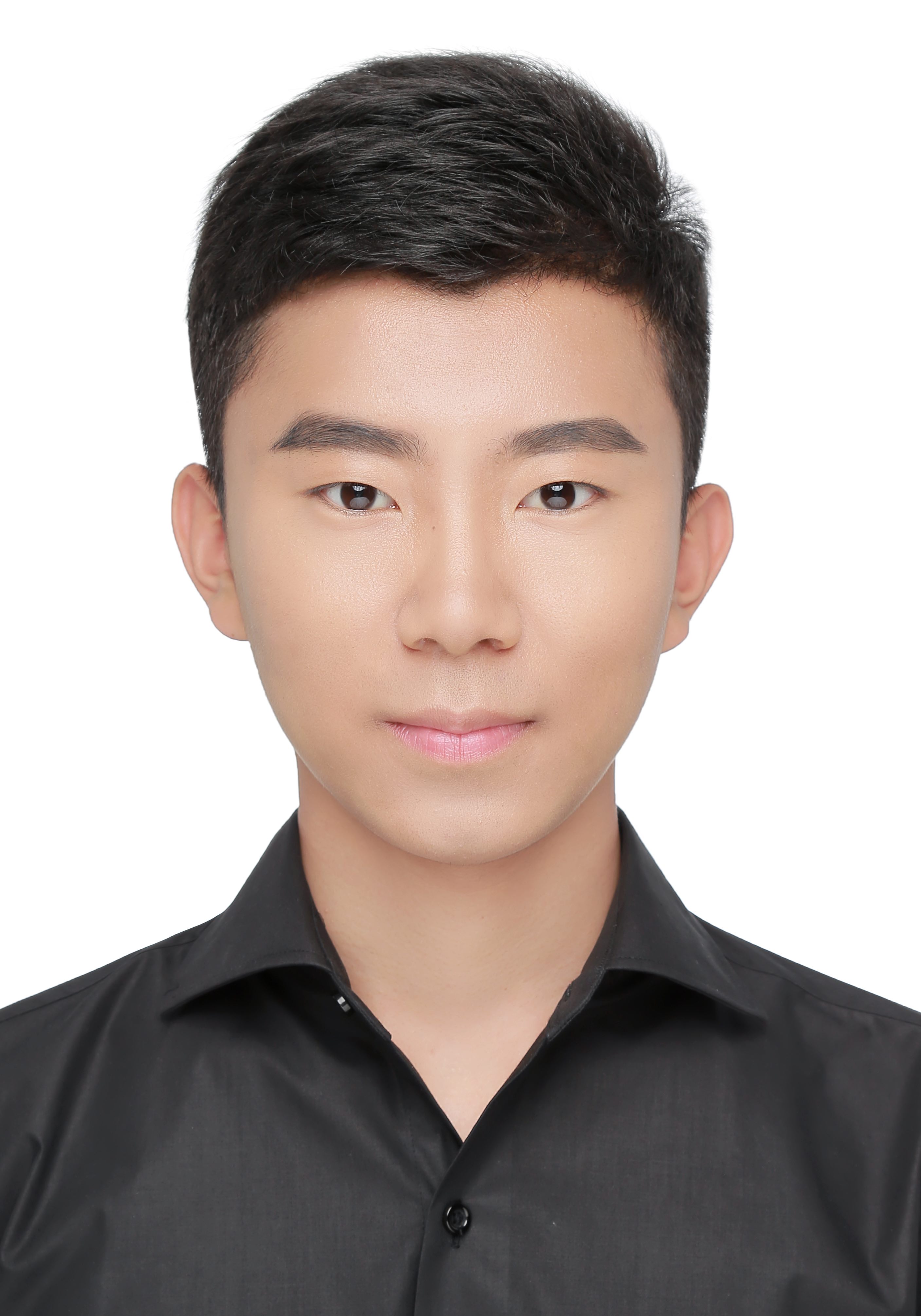}{Yurui Qiu}{is currently an algorithm engineer at Meituan. He received his Master's degree in Information and Communication Engineering from Tianjin University, Tianjin, China, in 2019. He was a Visiting Scholar at the School of Computing, National University of Singapore, in 2017, where he conducted research with Prof. M. Kankanhalli. His current research interests include multi-modality understanding and agentic AI.}

\mybio{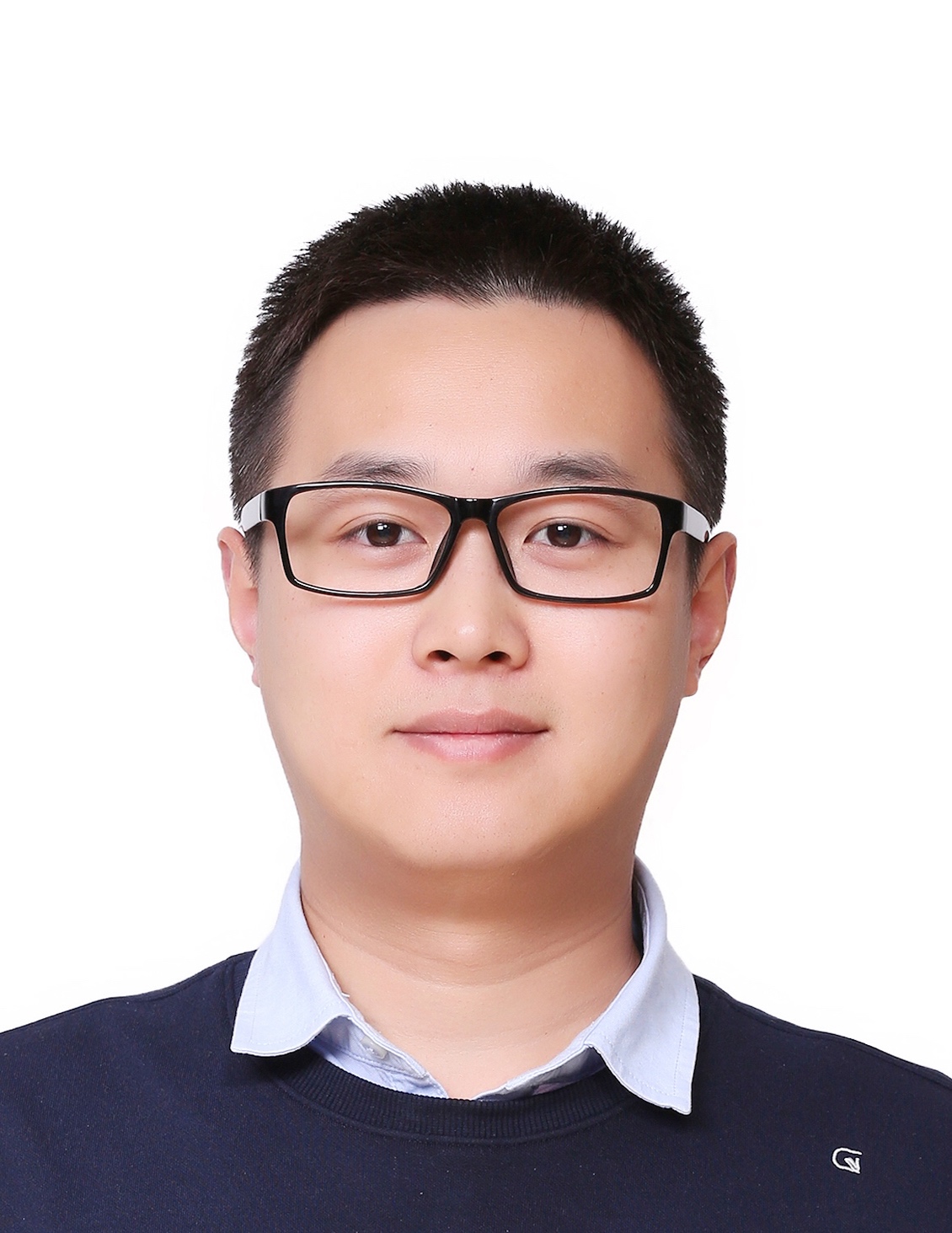}{Junfeng Luo}{is currently the leader of the Vision Agent Group at Meituan. He received his Ph.D. from Peking University in 2017 and has published several papers at conferences such as CVPR and AAAI. His research interests mainly include vision agent, agentic AI and vision understanding.}

\vspace{15mm}
\mybio{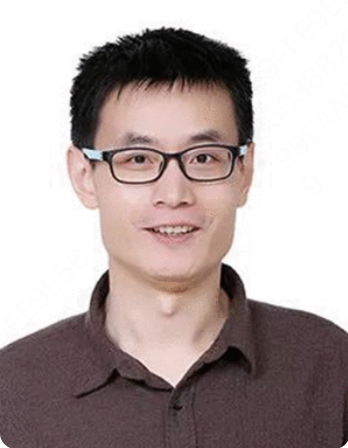}{Xiaoming Wei}{is currently the leader of Vision Intelligence Department at Meituan. His research interests focus on fine-grained image recognition, multimodal analysis and generation, etc. He has led the team and got top rankings in several fine-grained matches such as Herbarium 2022 FGVC9 (the 1st place), Product Recognition in CVPR2019 (the 2nd place). He has published 10+ papers in CVPR, ECCV, IJCAI, ACM MM, AAAI, etc.}

\vspace{15mm}
\mybio{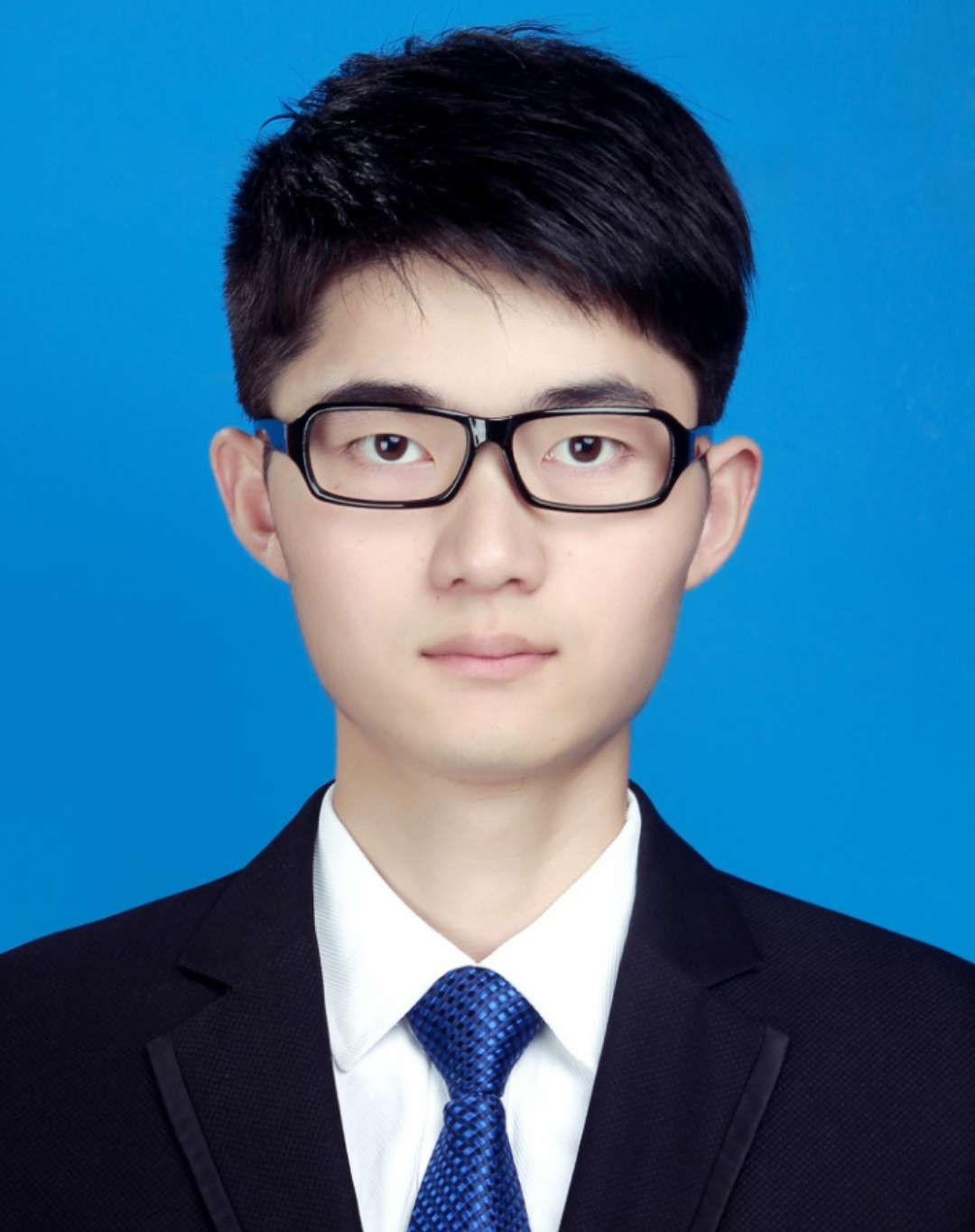}{Jialin Gao}{received the Ph.D. degree from School of Electronic Information and Electrical Engineering of Shanghai Jiao Tong University (SJTU), in 2022. Prior to that, he received the B.S. degree in electronic information engineering from the University of Electronic Science and Technology of China (UESTC), in 2016. He also has been a research fellow of AI Singapore, National University of Singapore and published multiple journal and conference papers in IEEE TCSVT, IEEE TMM, ICLR, EMNLP, SIGIR, AAAI, etc. His research interests include but are not limited to multimodal large language model,unified multimodal model, image generation, vision and language, and video understanding.}

\end{document}